\pdfoutput=1

\documentclass[11pt]{article}

\usepackage[final]{acl}

\usepackage{times}
\usepackage{latexsym}

\usepackage[T1]{fontenc}

\usepackage[utf8]{inputenc}

\usepackage{microtype}

\usepackage{inconsolata}

\usepackage{graphicx}

\usepackage{amsmath}
\usepackage[section]{placeins}
\usepackage{multirow}
\usepackage{makecell}
\usepackage{amssymb}
\usepackage{xspace}
\usepackage{enumitem}
\usepackage{tcolorbox}
\usepackage{booktabs}
\usepackage{multicol}
\usepackage{tabularx}
\usepackage{hyperref}
\usepackage{listings}
\usepackage{mathptmx}
\usepackage{xcolor}
\usepackage{subfigure}
\usepackage{color}
\usepackage[normalem]{ulem}
\usepackage{algorithm}
\usepackage{algorithmic}
\usepackage{float}
\usepackage{amssymb}
\usepackage{pifont}
\usepackage{orcidlink}

\usepackage[utf8]{inputenc}
\tcbuselibrary{skins, breakable}
\usepackage{marvosym}

\definecolor{headerblue}{RGB}{230, 240, 255} 
\definecolor{thoughtgray}{RGB}{248, 248, 248} 
\definecolor{toolblue}{RGB}{240, 248, 255}    
\definecolor{evidcream}{RGB}{255, 253, 245}   
\definecolor{scorehigh}{RGB}{220, 255, 220}  
\definecolor{scoremid}{RGB}{255, 248, 220}   
\definecolor{scorelow}{RGB}{255, 220, 220}   

\newtcolorbox[auto counter, number within=section]{promptbox}[1][]{ 
  breakable,
  enhanced,
  colback=gray!5,
  colframe=gray!50,
  coltitle=black,
  colbacktitle=gray!25,
  fonttitle=\small\bfseries\scshape,
  toptitle=3pt,
  bottomtitle=3pt,
  fontupper=\small\ttfamily,
  sharp corners,
  boxrule=0.5pt,
  left=5pt, right=5pt, top=5pt, bottom=5pt,
  title={Prompt~\thetcbcounter}, 
  label={prmt:\thetcbcounter},   
  #1 
}

\title{Instructions for *ACL Proceedings}

\author{First Author \\
  Affiliation / Address line 1 \\
  Affiliation / Address line 2 \\
  Affiliation / Address line 3 \\
  \texttt{email@domain} \\\And
  Second Author \\
  Affiliation / Address line 1 \\
  Affiliation / Address line 2 \\
  Affiliation / Address line 3 \\
  \texttt{email@domain} \\}

\title{AEScorer: An Agentic Evidence-Grounded Framework for\\ Graded Factuality Verification}

\author{
    Hui Huang\orcidlink{0000-0002-3544-9719}\textsuperscript{1,2}, 
    Muyun Yang\orcidlink{0000-0002-5940-0266}\textsuperscript{1},
    Yuki Arase\orcidlink{0000-0002-7271-6206}\textsuperscript{2} \\[5pt]
    \textsuperscript{1}Harbin Institute of Technology,
    \textsuperscript{2}Institute of Science Tokyo\\[3pt]
    \texttt{huanghui@stu.hit.edu.cn, arase@c.titech.ac.jp}
}

\begin{document}
\maketitle
\begin{abstract}

Despite the significant advancements of Large Language Models (LLMs), their factuality remains a critical challenge, creating a growing need for more nuanced factuality verification. Existing factuality verification methods do not capture graded judgments, even though factuality is better understood as a spectrum rather than a binary of right and wrong.
To bridge this gap, we focus on graded factuality verification and propose \textbf{AEScorer}, an agentic evidence-grounded framework with two stages: agentic evidence acquisition and graded scoring. AEScorer first gathers and refines external evidence through agentic search, and then predicts a scalar factuality score to distinguish nuanced differences in factual correctness.
We further construct \textbf{GradedVeriBench}, a benchmark for graded factuality verification spanning both general and multi-hop question answering. Experimental results on GradedVeriBench show that AEScorer substantially outperforms existing methods across both settings, demonstrating the value of coupling targeted evidence acquisition with graded scoring\footnote{Code and data are openly available at \url{https://github.com/HuihuiChyan/FactVeri-SFT}.}.



\end{abstract}

\section{Introduction}

Large Language Models (LLMs) such as GPT-4 have achieved remarkable progress in natural language generation and are now widely used in many applications \cite{zhao2023survey,liu2025comprehensive}. However, they can still generate plausible-sounding but factually incorrect content \cite{wang2023survey}. This phenomenon, commonly referred to as factual hallucination, remains a major barrier to deploying LLMs in scenarios where factual errors have practical consequences.

\begin{figure}[t]
    \centering
        \includegraphics[width=0.95\linewidth]{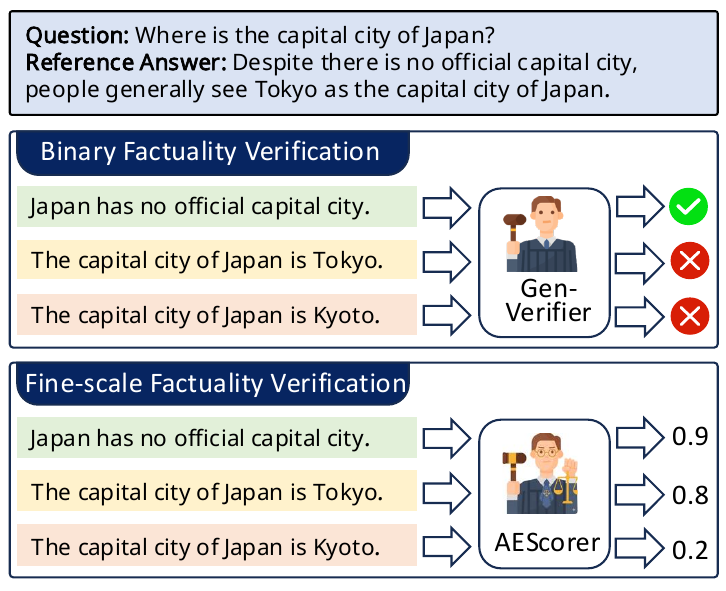}
        \caption{Illustrative comparison between coarse factuality verification and graded factuality verification. A single atomic fact may contain different degrees of factual error that binary verification cannot distinguish.}
    \label{figure:illustration}
\end{figure}


Existing factuality verification methods mostly operate in coarse categories such as correct or incorrect \cite{lin2022truthfulqa}. 
Recent work has discussed \emph{fine-grained} factuality verification focusing on the granularity of texts to verify and employed a segmented atomic claim as an evaluation unit, rather than a response \cite{min2023factscore,mitra2025factlens}.  
However, a short-form response or a single atomic fact can be factually correct to different degrees, as illustrated in Figure \ref{figure:illustration}.

We refer to this problem as \emph{graded factuality verification}: the task of distinguishing different degrees of factual incorrectness within a single atomic fact. This task is important in preference alignment applications, where reward models and reranking pipelines benefit from dense and informative factuality signals. It is also theoretically well motivated, as epistemological perspectives characterize truth as graded belief rather than a strict binary switch \cite{ramsey1926truth}, while fuzzy logic formalizes truth values on a continuous interval \cite{zadeh1965fuzzy}. These perspectives support treating factuality verification as a graded assessment problem rather than only a binary decision problem.

To address this task, we propose \textbf{AEScorer}, an agentic evidence-grounded framework for graded factuality verification. AEScorer first performs \emph{agentic evidence acquisition}, where the model interactively retrieves and refines external evidence for the target answer. It then conducts \emph{graded scoring}, predicting a scalar score that reflects nuanced factual differences rather than only coarse labels. AEScorer conditions its scoring on evidence retrieved through agentic search, enabling evidence-supported graded factuality scores at inference time. 
By training the scoring function using pairwise preferences, AEScorer avoids the need for absolute score annotations \cite{kasa-etal-2025-generative}.


We further introduce \textbf{GradedVeriBench}, a benchmark designed for graded factuality verification targeted at short-form question answering. GradedVeriBench is constructed through a ranking-based annotation pipeline that combines LLM-as-a-Judge \cite{calderon-etal-2025-alternative} with human verification to ensure reliable nuanced factuality judgments. Experimental results show that AEScorer substantially outperforms existing verification methods on GradedVeriBench and transfers to various verification scenarios.

Our primary contributions are threefold:
\begin{enumerate}[leftmargin=4mm, itemsep=1mm, parsep=0pt] 
    \item We formalize \emph{graded factuality verification}, which aims to distinguish different degrees of factual incorrectness within a single fact.
    \item We propose \textbf{AEScorer}, an agentic evidence-grounded framework that couples evidence acquisition with graded scoring.
    \item We introduce \textbf{GradedVeriBench}, the first benchmark for graded factuality verification. Experiment results on GradedVeriBench demonstrate that AEScorer is superior to existing methods in terms of graded verification.
\end{enumerate}


\section{Related Work}

\subsection{Factuality Verification} 

In the context of LLMs, factuality verification is the process of evaluating whether a piece of model-generated text, such as a claim or response, is factually accurate \cite{wang2023survey}. It also serves as a central mechanism for hallucination detection \cite{Huang_2025}.

Reliable benchmarks are fundamental for measuring factuality verification progress. Early works such as TruthfulQA \cite{lin2022truthfulqameasuringmodelsmimic} and HaluEval \cite{li-etal-2023-halueval} treat factuality mainly as a binary problem, providing foundational datasets for evaluating whether LLMs can recognize hallucinations and assess overall factual correctness. However, coarse labels are often insufficient for applications that require nuanced factuality feedback.

To address the complexity of longer generations, subsequent research has shifted toward fine-grained assessment. FactScore \cite{min2023factscore} popularized the \emph{Decompose-Then-Verify} paradigm, which decomposes long-form responses into atomic facts for individual verification. Follow-up works, including FELM \cite{chen2023felmbenchmarkingfactualityevaluation} and WiCE \cite{kamoi-etal-2023-wice}, further refined fine-grained verification over decomposed claims. 
Although this line of work has refined the \emph{unit of evaluation}, the resulting labels for each unit often remain binary or coarse. A graded benchmark that supports nuanced scoring within a single atomic fact remains absent.


\subsection{LLM-as-a-Verifier} 

LLM-based factuality verifiers can be broadly grouped into two technical paradigms: generative verifiers and discriminative verifiers. 


Generative verifiers leverage the generation ability of LLMs to retrieve external knowledge, produce reasoning traces, and output a final factuality assessment, represented by work such as FactScore and VeriScore \cite{song2024veriscore}. Recent methods have extended this paradigm in different directions: FGLR \cite{stacey2023atomic} enhances NLI-based reasoning by generating auxiliary premise facts, while FineSumFact \cite{oh2024learning} uses fine-grained LLM feedback to supervise factuality in summarization. 
However, prompting generative models to output numerical scores does not directly optimize them for scoring, leading to poorly calibrated graded scores.

In contrast, discriminative verifiers assess factuality more directly by mapping the input to a scalar factuality score or class label. For instance, \citet{paudel2025hallucinothallucinationdetectioncontext} use a classification head to validate responses against context and common knowledge. Similarly, \citet{hhem-2.1-open} propose a lightweight classifier for factual consistency evaluation. This formulation enables fast and efficient optimization for fine-grained scoring \cite{wang-etal-2024-f2rl}. However, without an explicit mechanism for acquiring external evidence, discriminative verifiers are often constrained by knowledge insufficiency.






\section{Proposed Method: AEScorer}

\begin{figure*}[t]
    \centering
        \includegraphics[width=1.0\linewidth]{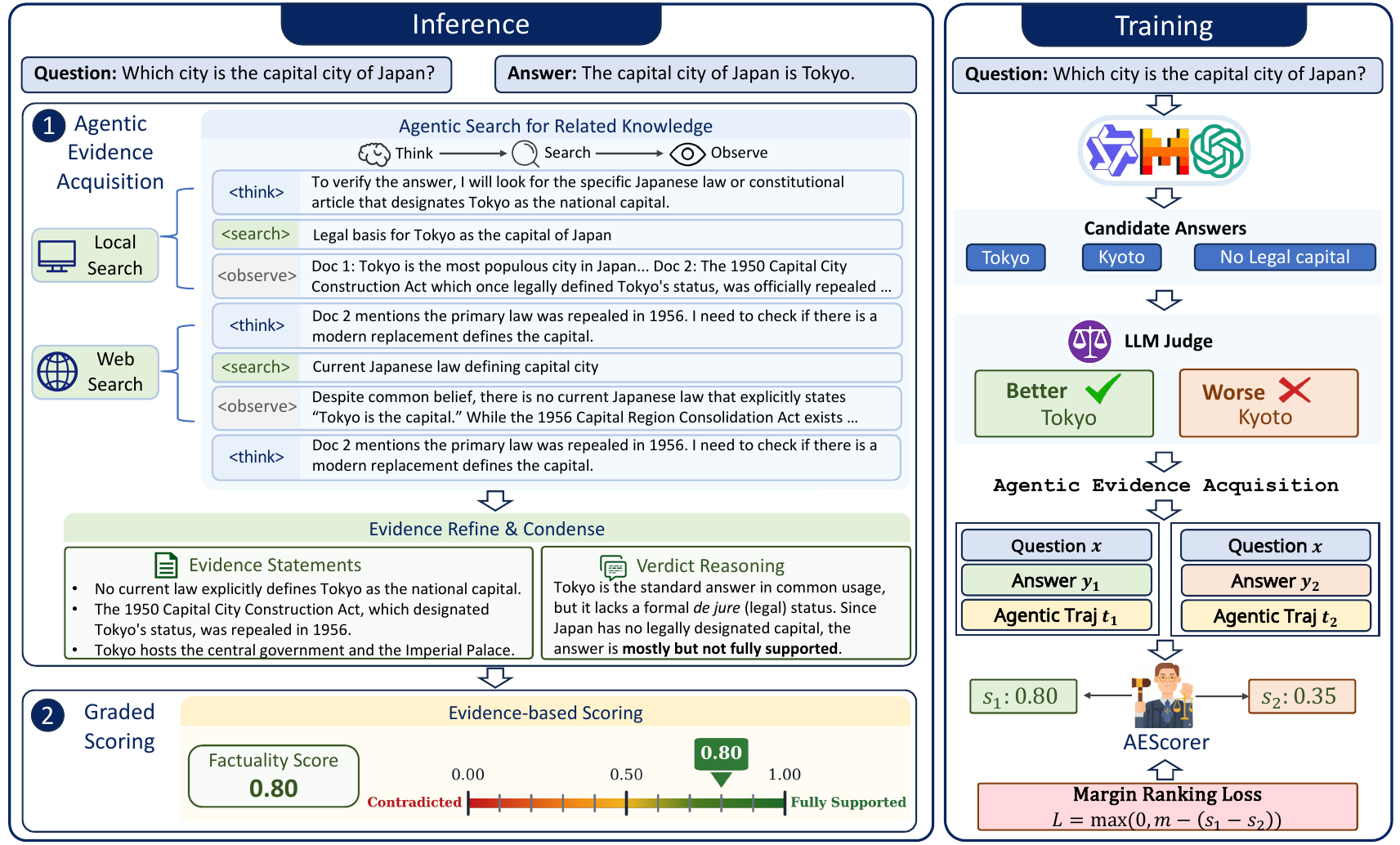}
        \caption{The overall pipeline of graded factuality verification with AEScorer. The left side illustrates unified inference with agentic evidence acquisition and graded scoring, while the right side shows pairwise data construction and the training process for its evidence-based scoring capability.}
    \label{figure:architecture}
\end{figure*}


Graded factuality verification is challenging because it requires both sufficient evidence grounding and sensitive score calibration. To address these two requirements, we propose AEScorer, which treats evidence acquisition and scoring as two tightly connected stages. As illustrated in Figure \ref{figure:architecture}, it first performs \emph{Agentic Evidence Acquisition} to collect and refine verification-oriented evidence, and then applies \emph{Graded Scoring} to produce a scalar factuality score grounded on the evidence.

\subsection{Agentic Evidence Acquisition}
\label{sec:architecture}


\begin{figure*}[t]
    \centering
        \includegraphics[width=1.0\linewidth]{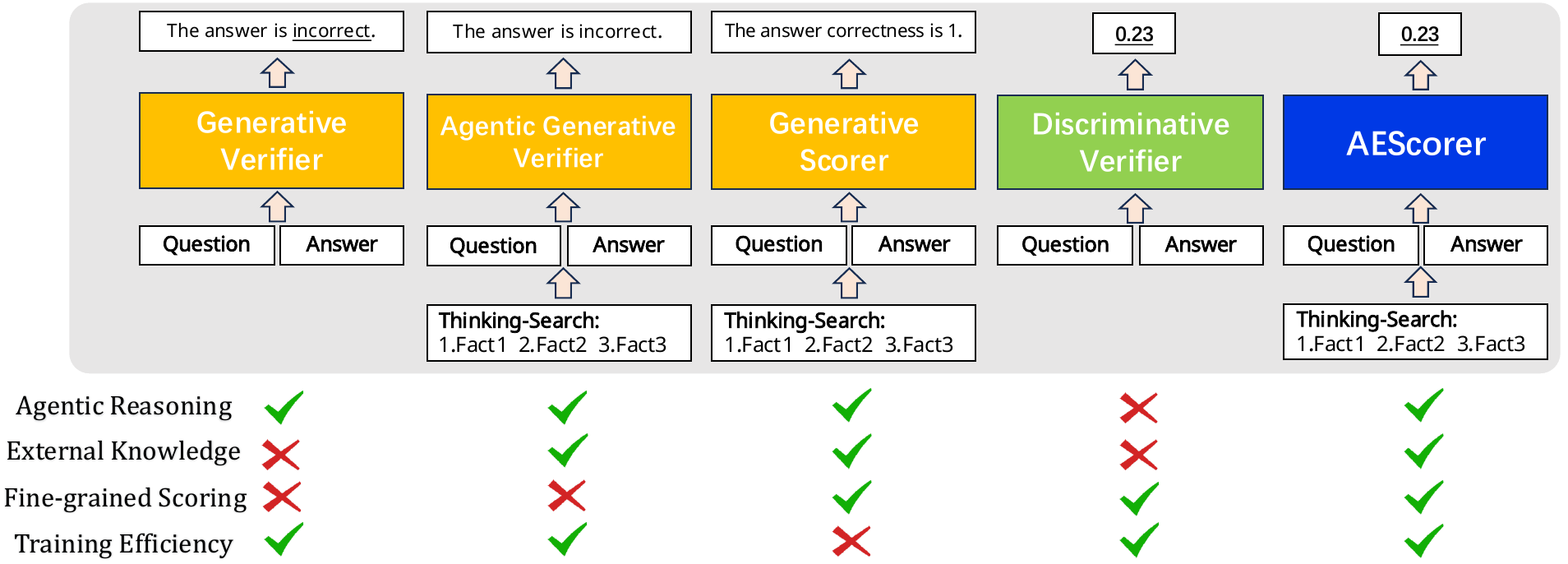}
        \caption{Comparison of different verification paradigms. We compare them in terms of agentic reasoning ability, accessibility to external knowledge, graded scoring ability, and training efficiency.}
    \label{figure:comparison}
\end{figure*}

\paragraph{Agentic Search}

Without access to external information, a verifier may suffer from severe knowledge gaps, especially when the target answer concerns temporal, entity-centric, or multi-hop facts. Therefore, AEScorer employs an agentic search mechanism \cite{jin2025searchr} that allows the model to plan evidence-seeking actions and autonomously retrieve evidence whenever its internal knowledge is insufficient for verification.

The retrieval process follows an iterative loop with the following steps:

\vspace{-2mm}
\begin{enumerate}[leftmargin=4mm, itemsep=1mm, parsep=0pt]
    \item \textbf{Think:} AEScorer plans the next verification action and formulates a targeted search query based on the missing evidence.
    \item \textbf{Search:} AEScorer retrieves information from external tools, including \texttt{WebSearch} for broad and up-to-date online evidence and \texttt{LocalSearch} for a trusted document repository.
    \item \textbf{Observe:} The retrieved results are incorporated into the running context to support the next round of reasoning. The loop terminates once AEScorer determines that the available evidence is sufficient for scoring.
\end{enumerate}
\vspace{-2mm}

This process produces an evidence-rich trajectory containing both retrieved knowledge and intermediate reasoning, establishing the factual grounding required for graded verification. 
In addition, the iterative loop allows AEScorer to adapt its search to the specific missing facts in the target answer.

\paragraph{Evidence Refinement}

Raw search trajectories are often verbose, noisy, and redundant. Since LLM-based systems operate under a finite attention budget and suffer from context decay \cite{mei2025survey}, AEScorer refines the acquired evidence before scoring. Concretely, it compresses the search trajectory into two compact components:

\vspace{-2mm}
\begin{enumerate} [itemsep=1mm, parsep=0pt]
    \item A list of useful evidence statements extracted from retrieved content;
    \item The reasoning traces necessary for the final verification decision.
\end{enumerate}
\vspace{-2mm}

By supplying the scorer with concise, high-signal input, we prevent the scoring stage from being dominated by irrelevant trajectory details.

\subsection{Graded Scoring}

Given the refined evidence, AEScorer predicts a scalar factuality score for the target answer. Unlike coarse verification, the objective here is not merely to classify an answer as correct or incorrect, but to distinguish nuanced differences in factual correctness for a single atomic fact or short response. To support such nuanced judgments, we implement the scoring component as a regression-style head over the model representation, allowing the framework to output continuous values rather than vocabulary-constrained discrete labels \cite{touvron2023llamaopenefficientfoundation}. This design makes the final decision directly optimized for scoring rather than dependent on the calibration of generated text.


While our goal is to predict a standalone scalar factuality score for each answer at inference time, training on absolute scores is challenging because human annotation of absolute factuality scores is difficult and can result in inconsistency. Therefore, we train AEScorer with pairwise preferences, which provide more reliable supervision for learning the scoring function as described in Section~\ref{sec:disc-training}.



\subsection{Scoring Optimization}
\label{sec:disc-training}

We construct training triplets of (\textit{question}, \textit{preferred answer}, \textit{dispreferred answer}), as shown on the right side of Figure \ref{figure:architecture}. We optimize the scoring component with a margin ranking loss:

\vspace{-2mm}
\begin{equation}
    L = \max\left(0, m - (f(x, y_+, t_+) - f(x, y_-, t_-))\right)
\end{equation}
\vspace{-2mm}

\noindent Here, $f(x, y, t)$ denotes the score assigned by AEScorer to answer $y$ for question $x$ given refined evidence trajectory $t$, $y_+$ is the preferred (more factual) answer, $y_-$ is the dispreferred (less factual) answer, $t_+$ and $t_-$ denote the refined evidence trajectories for the two answers obtained from the first stage, and $m$ is the ranking margin.

\paragraph{Pairwise Data Construction}


Directly annotating absolute scalar factuality scores for training is costly and often inconsistent. Therefore, we collect pairwise comparisons between candidate answers, which provide more reliable supervision while still enabling AEScorer to learn a standalone scoring function.


\vspace{-2mm}
\begin{enumerate}[leftmargin=4mm, itemsep=1mm, parsep=0pt]
    \item \textbf{Answer Generation:} Generate multiple diverse candidate answers for each question.
    \item \textbf{Initial Assessment:} Use an advanced LLM-as-a-judge to evaluate each answer against a reference, assigning an initial quality label.
    \item \textbf{Pair Sampling:} Randomly sample pairs consisting of a preferred (more factual) and a dispreferred (less factual) answer based on the labels.
    \item \textbf{Preference Verification:} Re-check the pairwise factuality order with the judge model.
    \item \textbf{Trajectory Construction:} Run the agentic evidence acquisition stage for each answer and store its refined evidence trajectory.
\end{enumerate}
\vspace{-2mm}


By training on this dataset with the margin ranking loss, AEScorer learns to assign higher scores to answers that are more factual with respect to reference information. Consequently, during inference, the trained verifier can predict a scalar score for a single, standalone answer, although supervision is provided through pairwise preferences\footnote{Notice this score is a discriminative relative-ranking signal rather than a probabilistically calibrated absolute metric.}.

\paragraph{Parameter Efficiency}


To ensure practical applicability and minimize storage overhead, we employ Low-Rank Adaptation (LoRA) to train the scoring capability \cite{hu2021loralowrankadaptationlarge}. This parameter-efficient fine-tuning strategy ensures that the storage footprint of the final scorer is only marginally larger than that of the base generative model. Furthermore, it enhances the method’s adaptability across multi-domain scenarios \cite{tian-etal-2025-adapters}.


\subsection{Comparison to Existing Paradigms}
\label{section:comparison}



As summarized in Figure \ref{figure:comparison}, existing verifiers have distinct limitations as follows.

\vspace{-2mm}
\begin{enumerate}[leftmargin=4mm, itemsep=1mm, parsep=0pt]
    \item \textbf{Naive Generative Verifier} relies solely on LLMs' internal knowledge and fails when claims exceed the model's knowledge boundary.
    \item \textbf{Agentic-Generative Verifier} such as \citet{jin2025searchr}, has access to external knowledge but can not generate fine-grained scores.
    \item \textbf{Agentic-Generative Scorer} such as \citet{li2023revisit}, is constrained by scalar score generation ability and the lack of absolute scoring data for effective optimization.
    \item \textbf{Naive Discriminative Verifier} such as \citet{liu-etal-2024-hit}, is easy to be optimized for scoring with a pairwise objective but lacks external knowledge.
\end{enumerate}
\vspace{-2mm}


AEScorer addresses these shortcomings by embodying the idea of \emph{agentic search as feature engineering}. It combines interactive evidence gathering with a ranking-trained scoring stage, enabling both efficient training and evidence-based scoring. 


\begin{figure}[t]
    \centering
        \includegraphics[width=1.0\linewidth]{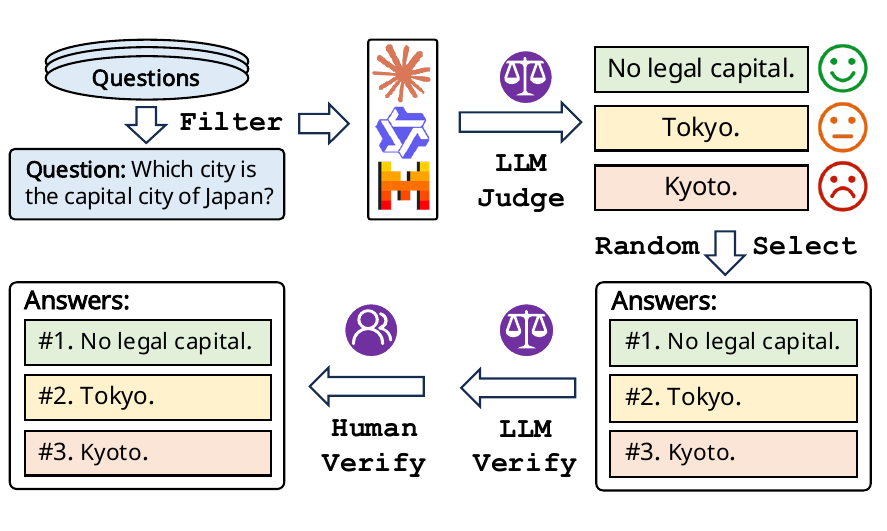}
        \caption{Construction process of GradedVeriBench.}
    \label{figure:data-construction}
\end{figure}

\section{GradedVeriBench}

\begin{figure}[t]
    \centering
        \includegraphics[width=1.0\linewidth]{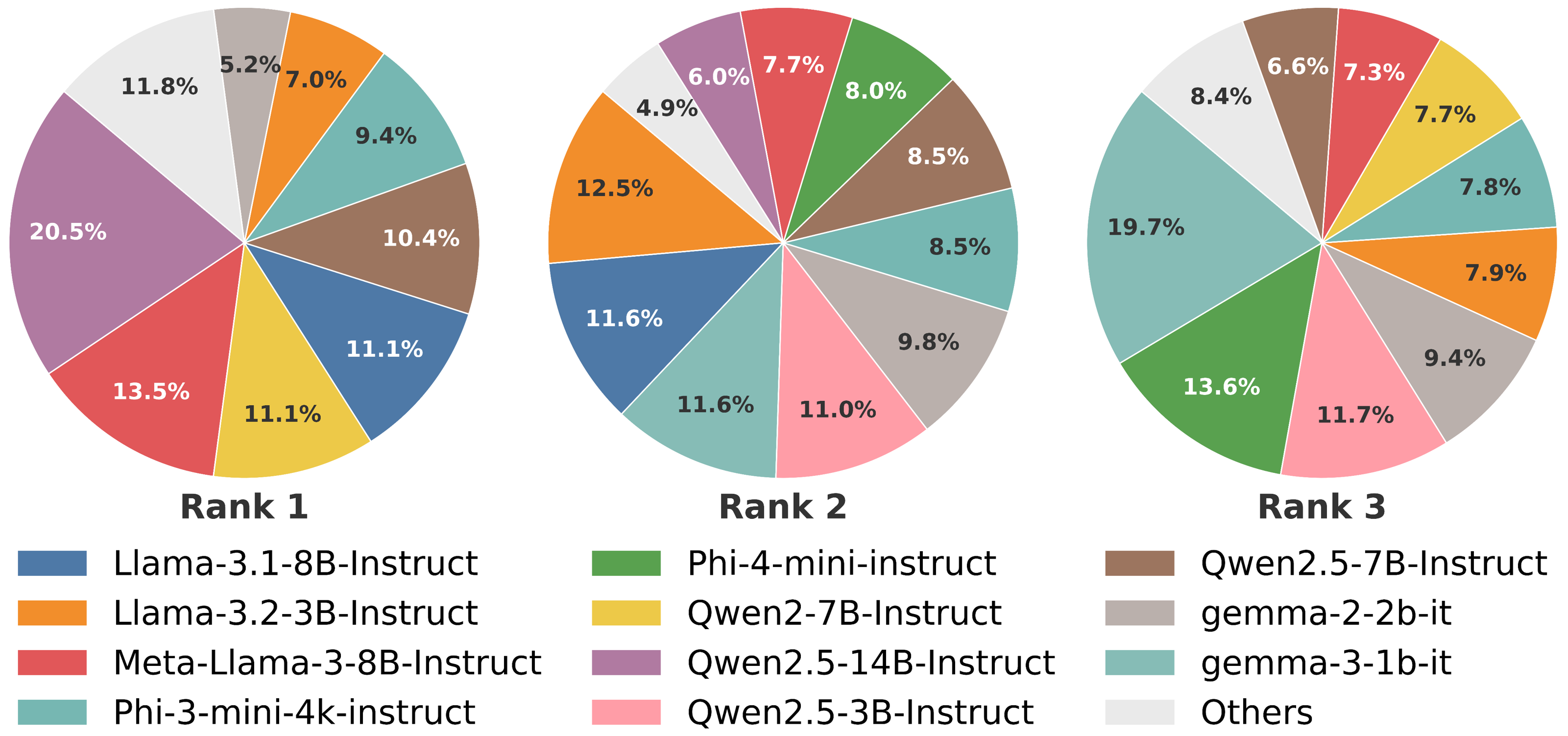}
        \caption{Model distribution of different answer ranks in GradedVeriBench. Rank 1 has the best factuality.}
    \label{fig:model-dist}
\end{figure}

\begin{table}[t]
    \centering
    \resizebox{0.48 \textwidth}{!}{
    \begin{tabular}{cccccc}
    \hline
    \multicolumn{2}{c}{\textbf{NQ}}       & \multicolumn{2}{c}{\textbf{TriviaQA}} & \multicolumn{2}{c}{\textbf{PopQA}} \\ \hline
    Number            & Length            & Number            & Length            & Number           & Length          \\
    $203$               & $15.7$             & $372$               & $13.7$             & $365$              & $14.3$           \\ \hline
    \multicolumn{2}{c}{\textbf{HotpotQA}} & \multicolumn{2}{c}{\textbf{Musique}}  & \multicolumn{2}{c}{\textbf{2wiki}} \\ \hline
    Number            & Length            & Number            & Length            & Number           & Length          \\
    $287$               & $16.1$             & $154$               & $21.0$             & $298$              & $14.9$           \\ \hline
    \end{tabular}}
    \caption{Data statistics of GradedVeriBench.}
    \label{tab:data-statistics}
\end{table}

Existing factuality benchmarks typically focus on either coarse verification or long-form claim decomposition. To address the missing setting of \emph{graded factuality verification}, we introduce GradedVeriBench, a benchmark designed for evaluating relative factual differences among short-form answers, as shown in Figure~\ref{figure:data-construction}. 

Since determining the relative order of responses is significantly more reliable than assigning absolute pointwise scores, GradedVeriBench adopts a ranking-based annotation framework to improve data quality. This design converts nuanced factuality evaluation into a controlled ranking problem. The construction process is as follows:

\vspace{-2mm}
\begin{enumerate}[leftmargin=4mm, itemsep=1mm, parsep=0pt]
    \item \textbf{Instruction Filtering}: Select only verifiable questions that possess a single, unambiguous correct answer to ensure objective evaluation.
    \item \textbf{Answer Pool Generation}: Leverage representative LLMs to generate a diverse pool of candidate answers for each question.
    \item \textbf{Initial Verification}: Leverage a proprietary judge to assign an initial category to each answer: \texttt{Correct}, \texttt{Incorrect} or \texttt{Intermediate}.
    \item \textbf{Random Selection}: Randomly sample three answers for each question with one from each of \texttt{correct}, \texttt{intermediate} and \texttt{incorrect} ones.
    \item \textbf{LLM\&Human-based Verification}: Leverage both an LLM and human evaluators to verify the relative order of factuality from best to worst among the three sampled answers. Three graduate-student annotators independently judge whether each LLM-proposed ranking is valid.
\end{enumerate}
\vspace{-2mm}

\begin{table*}[!t]
\centering
\resizebox{0.92 \textwidth}{!}{
\begin{tabular}{ccccccccccccc}
\hline
\multirow{3}{*}{\textbf{Method}} & \multicolumn{4}{c}{\multirow{2}{*}{\textbf{Features}}}                                                                                                & \multicolumn{6}{c}{\textbf{General QA Verification}}                                                         & \multicolumn{2}{c}{\multirow{2}{*}{\textbf{Average}}} \\ \cline{6-11}
                                 & \multicolumn{4}{c}{}                                                                                                                                  & \multicolumn{2}{c}{\textbf{NQ}} & \multicolumn{2}{c}{\textbf{TriviaQA}} & \multicolumn{2}{c}{\textbf{PopQA}} & \multicolumn{2}{c}{}                                  \\ \cline{2-13} 
                                 & \textbf{AR}                         & \textbf{EK}                         & \textbf{GS}                         & \textbf{TE}                         & \textbf{P@1}   & \textbf{K-Tau} & \textbf{P@1}      & \textbf{K-Tau}    & \textbf{P@1}     & \textbf{K-Tau}  & \textbf{P@1}              & \textbf{K-Tau}            \\ \hline
\multicolumn{5}{c}{MiniCheck w/ Flan-T5 (\citeauthor{tang2024minicheck})}                                                                                                                                                 & 67.98          & 59.28          & 68.82             & 62.19             & 56.99            & 47.40           & 64.60                     & 56.29                     \\
\multicolumn{5}{c}{MiniCheck w/ MiniCheck-7B (\citeauthor{tang2024minicheck})}                                                                                                                                            & 71.92          & 66.50          & 74.46             & 69.89             & 67.40            & 57.44           & 71.26                     & 64.61                     \\
\multicolumn{5}{c}{HHEM-2.1-Open (\citeauthor{hhem-2.1-open})}                                                                                                                                                        & 66.01          & 56.98          & 69.35             & 60.04             & 57.53            & 45.57           & 64.30                     & 54.20                     \\
\multicolumn{5}{c}{FactScore w/ GPT-4 (\citeauthor{min2023factscore})}                                                                                                                                                   & 77.64          & 70.74          & 70.00             & 70.11             & 70.14            & 61.32           & 72.59                     & 67.39                     \\
\multicolumn{5}{c}{VeriScore w/ GPT-4o (\citeauthor{song2024veriscore})}                                                                                                                                                 & 78.46          & 79.68          & 84.41             & 81.81             & 78.57            & 76.67           & 80.48                     & 79.39                     \\ \hline
\multicolumn{13}{l}{\textit{Results on Qwen-2.5-7B-Instruct.}}                                                                                                                                                                                                                                                                                                  \\ \hline
Generative Verifier              & \ding{51}          & \ding{55}          & \ding{55}          & \ding{51}          & 71.92          & 62.56          & 69.81             & 61.37             & 71.15            & 58.42           & 70.96                     & 60.78                     \\
Agentic-Generative Verifier      & \ding{51}          & \ding{51}          & \ding{55}          & \ding{51}          & 77.83          & 71.43          & 80.91             & 74.73             & 81.10            & 76.26           & 79.95                     & 74.14                     \\
Agentic-Generative Scorer        & \ding{51}          & \ding{51}          & \ding{51}          & \ding{55}          & 79.31          & 75.70          & 78.98             & 72.69             & 81.37            & 73.33           & 79.89                     & 73.91                     \\
Discriminative Verifier          & \ding{55}          & \ding{55}          & \ding{51}          & \ding{51}          & 67.49          & 56.32          & 75.81             & 63.80             & 73.42            & 61.10           & 72.24                     & 60.41                     \\ \hline
\textbf{AEScorer}                    & \textbf{\ding{51}} & \textbf{\ding{51}} & \textbf{\ding{51}} & \textbf{\ding{51}} & \textbf{84.24} & \textbf{82.92} & \textbf{90.86}    & \textbf{88.53}    & \textbf{90.14}   & \textbf{85.21}  & \textbf{88.41}            & \textbf{85.55}            \\
w/o compression                  & \ding{51}          & \ding{51}          & \ding{51}          & \ding{51}          & 81.28          & 77.67          & 83.87             & 79.39             & 83.56            & 79.91           & 82.90                     & 78.99                     \\ \hline
\multicolumn{13}{l}{\textit{Results on Qwen-3-4B-Instruct.}}                                                                                                                                                                                                                                                                                                    \\ \hline
Generative Verifier              & \ding{51}          & \ding{55}          & \ding{55}          & \ding{51}          & 71.57          & 63.45          & 70.19             & 58.96             & 68.07            & 57.24           & 69.94                     & 59.88                     \\
Agentic-Generative Verifier      & \ding{51}          & \ding{51}          & \ding{55}          & \ding{51}          & 77.34          & 71.10          & 81.99             & 76.16             & 82.47            & 76.99           & 80.60                     & 74.75                     \\
Agentic-Generative Scorer        & \ding{51}          & \ding{51}          & \ding{51}          & \ding{55}          & 86.21          & 79.31          & 78.98             & 72.69             & 81.37            & 73.33           & 82.19                     & 75.11                     \\
Discriminative Verifier          & \ding{55}          & \ding{55}          & \ding{51}          & \ding{51}          & 67.49          & 55.01          & 72.04             & 54.84             & 66.85            & 52.51           & 68.79                     & 54.12                     \\ \hline
\textbf{AEScorer}                    & \textbf{\ding{51}} & \textbf{\ding{51}} & \textbf{\ding{51}} & \textbf{\ding{51}} & \textbf{85.22} & \textbf{81.28} & \textbf{88.98}    & \textbf{84.23}    & \textbf{88.77}   & \textbf{83.20}  & \textbf{87.66}            & \textbf{82.90}            \\
w/o compression                  & \ding{51}          & \ding{51}          & \ding{51}          & \ding{51}          & 79.31          & 78.98          & 87.90             & 83.69             & 87.67            & 82.10           & 84.96                     & 81.59                     \\ \hline
\end{tabular}}
\caption{Experimental results of different verifiers on general QA verification (AR: Agentic Reasoning, EK: External Knowledge, GS: Graded Scoring, TE: Training Efficiency)}
\label{tab:main01}
\end{table*}

\begin{table*}[!h]
\centering
\resizebox{0.92 \textwidth}{!}{
\begin{tabular}{ccccccccccccc}
\hline
\multirow{3}{*}{\textbf{Method}} & \multicolumn{4}{c}{\multirow{2}{*}{\textbf{Features}}}                                                                                                & \multicolumn{6}{c}{\textbf{Multi-Hop QA Verification}}                                                            & \multicolumn{2}{c}{\multirow{2}{*}{\textbf{Average}}} \\ \cline{6-11}
                                 & \multicolumn{4}{c}{}                                                                                                                                  & \multicolumn{2}{c}{\textbf{HotpotQA}} & \multicolumn{2}{c}{\textbf{Musique}} & \multicolumn{2}{c}{\textbf{2wiki}} & \multicolumn{2}{c}{}                                  \\ \cline{2-13} 
                                 & \textbf{AR}                         & \textbf{EK}                         & \textbf{GS}                         & \textbf{TE}                         & \textbf{P@1}      & \textbf{K-Tau}    & \textbf{P@1}      & \textbf{K-Tau}   & \textbf{P@1}     & \textbf{K-Tau}  & \textbf{P@1}              & \textbf{K-Tau}            \\ \hline
\multicolumn{5}{c}{MiniCheck w/ Flan-T5 (\citeauthor{tang2024minicheck})}                                                                                                                                                 & 56.99             & 47.40             & 46.10             & 41.56            & 51.34            & 48.10           & 51.48                     & 45.69                     \\
\multicolumn{5}{c}{MiniCheck w/ MiniCheck-7B (\citeauthor{tang2024minicheck})}                                                                                                                                            & 62.72             & 50.99             & 53.25             & 46.75            & 58.05            & 53.69           & 58.01                     & 50.48                     \\
\multicolumn{5}{c}{HHEM-2.1-Open (\citeauthor{hhem-2.1-open})}                                                                                                                                                        & 52.61             & 33.80             & 50.65             & 46.32            & 58.72            & 47.43           & 53.99                     & 42.52                     \\
\multicolumn{5}{c}{FactScore w/ GPT-4 (\citeauthor{min2023factscore})}                                                                                                                                                   & 73.21             & 54.51             & 72.34             & 69.78            & 68.95            & 62.46           & 71.50                     & 62.25                     \\
\multicolumn{5}{c}{VeriScore w/ GPT-4o (\citeauthor{song2024veriscore})}                                                                                                                                                   & 78.66             & 72.63            & 73.23             & 70.54            & 69.69            & 67.82           & 73.86                     & 70.33                     \\ \hline
\multicolumn{13}{l}{\textit{Results on Qwen-2.5-7B-Instruct.}}                                                                                                                                                                                                                                                                                                       \\ \hline
Generative Verifier              & \ding{51}          & \ding{55}          & \ding{55}          & \ding{51}          & 63.99             & 41.96             & 57.14             & 48.92            & 48.15            & 24.13           & 56.43                     & 38.34                     \\
Agentic-Generative Verifier      & \ding{51}          & \ding{51}          & \ding{55}          & \ding{51}          & 72.13             & 60.51             & 57.79             & 53.25            & 67.11            & 49.22           & 65.68                     & 54.33                     \\
Agentic-Generative Scorer        & \ding{51}          & \ding{51}          & \ding{51}          & \ding{55}          & 75.26             & 59.12             & 73.38             & 61.90            & 66.11            & 42.73           & 71.58                     & 54.58                     \\
Discriminative Verifier          & \ding{55}          & \ding{55}          & \ding{51}          & \ding{51}          & 63.76             & 45.18             & 64.29             & 51.95            & 59.40            & 32.89           & 62.48                     & 43.34                     \\ \hline
\textbf{AEScorer}                    & \textbf{\ding{51}} & \textbf{\ding{51}} & \textbf{\ding{51}} & \textbf{\ding{51}} & \textbf{83.28}    & \textbf{77.24}    & \textbf{79.87}    & \textbf{77.92}   & \textbf{76.51}   & \textbf{68.68}  & \textbf{79.89}            & \textbf{74.61}            \\
w/o compression                  & \ding{51}          & \ding{51}          & \ding{51}          & \ding{51}          & 72.82             & 64.46             & 65.58             & 58.01            & 65.10            & 51.68           & 67.83                     & 58.05                     \\ \hline
\multicolumn{13}{l}{\textit{Results on Qwen-3-4B-Instruct.}}                                                                                                                                                                                                                                                                                                         \\ \hline
Generative Verifier              & \ding{51}          & \ding{55}          & \ding{55}          & \ding{51}          & 55.39             & 41.76             & 59.85             & 50.36            & 52.07            & 31.72           & 55.77                     & 41.28                     \\
Agentic-Generative Verifier      & \ding{51}          & \ding{51}          & \ding{55}          & \ding{51}          & 78.05             & 66.78             & 66.23             & 58.44            & 71.14            & 50.56           & 71.81                     & 58.59                     \\
Agentic-Generative Scorer        & \ding{51}          & \ding{51}          & \ding{51}          & \ding{55}          & 80.28             & 69.95             & 68.18             & 65.37            & 72.20            & 55.71           & 73.55                     & 63.68                     \\
Discriminative Verifier          & \ding{55}          & \ding{55}          & \ding{51}          & \ding{51}          & 58.54             & 43.55             & 61.04             & 40.69            & 59.60            & 34.90           & 59.73                     & 39.71                     \\ \hline
\textbf{AEScorer}                    & \textbf{\ding{51}} & \textbf{\ding{51}} & \textbf{\ding{51}} & \textbf{\ding{51}} & \textbf{87.80}    & \textbf{80.49}    & \textbf{81.82}    & \textbf{73.59}   & \textbf{81.21}   & \textbf{69.35}  & \textbf{83.61}            & \textbf{74.48}            \\
w/o compression                  & \ding{51}          & \ding{51}          & \ding{51}          & \ding{51}          & 83.62             & 73.75             & 80.52             & 72.73            & 72.82            & 66.44           & 78.99                     & 70.97                     \\ \hline
\end{tabular}}
\caption{Experimental results of different verifiers on multi-hop QA verification.}
\label{tab:main02}
\end{table*}

Let $N$ denote the number of samples sent for human verification, we define agreement rate as

{
\vspace{-2mm}
\begin{equation}
\text{Agreement rate} = \frac{1}{N}\sum_{i=1}^{N}\mathbb{I}\!\left(\substack{\text{all three annotators}\\\text{give the same judgment}}\right).
\label{eq:human-agreement}
\end{equation}
\vspace{-2mm}
}

This quantity measures unanimity when validating the LLM-proposed rankings in the pre-filtering pool. On GradedVeriBench, the agreement rate among the annotators was 64.8\%, showing that nuanced factuality annotation is non-trivial.




GradedVeriBench is constructed from six question-answering (QA) datasets, categorized as follows: 1) General QA: NQ \cite{kwiatkowski2019natural}, TriviaQA \cite{joshi2017triviaqa}, and PopQA \cite{mallen2022not}. 2) Multi-hop QA: HotpotQA \cite{yang2018hotpotqa}, 2Wiki \cite{xanh2020_2wikimultihop}, and Musique~\cite{trivedi-etal-2022-musique}. This enables comprehensive evaluation of graded factuality verification in both general and reasoning-intensive scenarios. We also present data statistics in Figure \ref{fig:model-dist} and Table \ref{tab:data-statistics}, demonstrating diverse model sources and balanced domain coverage\footnote{Two data samples are presented in Appendix \ref{app:data-sample}.}. 


\paragraph{Evaluation Metrics}
Performance on this benchmark is evaluated using \textbf{Kendall's $\tau$} correlation coefficient to assess the global alignment with human rankings, and \textbf{Precision@1} to verify the model's ability to identify the most factual response.

\section{Experiments}

We conduct a comprehensive evaluation of AEScorer on GradedVeriBench, together with ablation experiments on various verification scenarios. 


\subsection{Settings}



Because there is no existing study that combines agentic search and graded factuality verification, we compare AEScorer with prevalent factuality verification methods as baselines: FactScore \cite{min2023factscore}, MiniCheck \cite{tang2024minicheck}, HHEM \cite{hhem-2.1-open}, and VeriScore \cite{song2024veriscore}. 

We also compare AEScorer with the methods discussed in Section \ref{section:comparison} as an ablation study. 
For training the scoring capability of AEScorer and the Discriminative Verifier, we merge the training sets of NQ and HotpotQA to form a corpus following \citet{jin2025searchr}, using the pipeline illustrated in Figure \ref{figure:architecture}.
Other verifiers directly use the generative ability of LLMs for verification.  
The base models are Qwen2.5-7B-Instruct and Qwen3-4B-Instruct \cite{yang2025qwen3}, both with strong capabilities in reasoning and tool use. 
For retrieval, we employ Google Search for \texttt{WebSearch} and Wikipedia dumps dated up to 2018 for \texttt{LocalSearch} (see Appendix \ref{app:implementations} for more details).

\subsection{Main Results}
\label{sec:main_results}

As shown in Tables \ref{tab:main01} and \ref{tab:main02}, the performance of both generative and discriminative verifiers drops significantly without the aid of agentic search. This is because the static, parametric knowledge within LLMs is limited and cannot be updated over time. This limitation is particularly pronounced given the relatively small size of the backbone models.

Moreover, based on the same agentic search process, AEScorer significantly outperforms its generative counterparts\footnote{A case study of the different verifiers is in Appendix \ref{app:case-study}.}. Although we adjust the prompts of generative baselines to score individual responses or directly output rankings, their generation-oriented objectives still hinder precise graded verification. In contrast, our optimized scorer trained with a ranking objective offers better separation between answers with nuanced factual correctness.

Furthermore, removing the evidence refinement step leads to a substantial drop in AEScorer's performance, particularly in multi-hop scenarios. This validates the role of trajectory refinement in preserving high-signal evidence for effective scoring.

Other competing methods also underperform because they either lack dedicated optimization for graded scoring or are primarily designed for long-form verification. Despite its comparatively small backbone size, AEScorer achieves the best performance thanks to its evidence-grounded scoring design, showing that task-specific supervision can compensate for model scale in this setting.

\subsection{Application to LLM Generation}
\setlength{\tabcolsep}{2pt}
\begin{table}[!t]
    \centering
    \resizebox{0.48 \textwidth}{!}{
        \begin{tabular}{ccccc}
        \hline
        \multirow{2}{*}{\textbf{Selector}}    & \multirow{2}{*}{\textbf{Model}}      & \multicolumn{3}{c}{\textbf{Token-level F1 Score}}         \\
                                              &                                      & \textbf{NQ\_Test} & \textbf{Musique} & \textbf{Bamboogle} \\ \hline
        \multicolumn{2}{c}{Meta-Llama-3-8B-Inst}                                     & $23.82$             & $11.14$            & $15.02$              \\
        FactScore                             & GPT-4                                & $25.81$             & $10.44$            & $19.05$              \\
        AEScorer                                  & Qwen-2.5-7B-Inst                     & $24.38$             & \textbf{14.60}   & \textbf{23.40}     \\
        AEScorer                                  & Qwen-3-4B-Inst                       & \textbf{26.14}    & $14.28$            & $21.39$              \\ \hline
        \multicolumn{2}{c}{Llama-3.1-8B-Inst}                                        & $19.22$             & $9.81$             & $13.72$              \\
        FactScore                             & GPT-4                                & $19.72$             & $11.36$            & $13.42$              \\
        AEScorer                                  & Qwen-2.5-7B-Inst                     & \textbf{22.23}    & \textbf{11.78}   & $17.58$              \\
        AEScorer                                  & Qwen-3-4B-Inst                       & $22.15$             & $11.61$            & \textbf{19.01}     \\ \hline
        \end{tabular}}
        \caption{Best-of-N ($N=16$) selection results on question answering tasks with different verifiers.}
    \label{tab:best-of-N}
\end{table}
\setlength{\tabcolsep}{4pt}

Graded verification is highly valuable for improving downstream generation factuality because it can rank candidate answers by factual quality rather than merely filter obviously incorrect ones. To demonstrate this, we use AEScorer for best-of-$N$ selection in generation. 
Specifically, we employ Meta-Llama-3-8B-Instruct and Llama-3.1-8B-Instruct to generate multiple candidate answers for three distinct QA datasets. 
The verifiers then select the most factual response from the candidates.

The F1 score on question answering test sets is shown in Table \ref{tab:best-of-N}. The results confirm that AEScorer improves factuality for LLM generation, demonstrating the usefulness of graded verification for reranking. In contrast, FactScore struggles to separate nuanced factual differences despite using a much larger model, likely because it was designed around long-form claim decomposition rather than graded short-answer factuality. 


\begin{table}[t]
\centering
\resizebox{0.48 \textwidth}{!}{
\begin{tabular}{cccccc}
\hline
\multirow{2}{*}{\textbf{Method}} & \multirow{2}{*}{\textbf{Model}} & \multicolumn{2}{c}{\textbf{TriviaQA}} & \multicolumn{2}{c}{\textbf{HotpotQA}} \\
                                 &                                 & \textbf{ACC}        & \textbf{F1}         & \textbf{ACC}        & \textbf{F1}         \\ \hline
FactScore                        & GPT-4                           & $59.95$             & $52.85$             & $61.85$             & $55.21$             \\ \hline
G-Verifier                       & Qwen2.5-7B-Inst                 & $66.14$             & $64.82$             & $43.53$             & $42.49$             \\
AG-Verifier                      & Qwen2.5-7B-Inst                 & $75.98$             & $74.89$             & $50.20$             & $49.76$             \\
AEScorer                             & Qwen2.5-7B-Inst                 & \textbf{88.58}      & \textbf{88.53}      & \textbf{70.98}      & \textbf{70.88}      \\ \hline
G-Verifier                       & Qwen3-4B-Inst                   & $66.38$             & $65.45$             & $45.29$             & $44.85$             \\
AG-Verifier                      & Qwen3-4B-Inst                   & $74.41$             & $73.66$             & $60.39$             & $60.51$             \\
AEScorer                      & Qwen3-4B-Inst                   & \textbf{88.19}    & \textbf{88.17}    & \textbf{72.94}    & \textbf{72.46}    \\ \hline
\end{tabular}}
\caption{Results on binary factuality verification}
\label{tab:binary}
\end{table}

\begin{table}[t]
\centering
\resizebox{0.48 \textwidth}{!}{
\begin{tabular}{cccc}
\hline
\textbf{Method} & \textbf{Model}      & \multicolumn{2}{c}{\textbf{FactScore Dataset}}       \\ 
\textbf{}       & \textbf{}           & \textbf{Precision@1} & \textbf{Kendall-tau} \\ \hline
FactScore       & GPT-4               & $60.62$       & $45.00$      \\ \hline
G-Verifier      & Qwen2.5-7B-Inst     & $33.12$                & $5.83$                \\
AG-Verifier     & Qwen2.5-7B-Inst     & $46.25$                & $25.83$               \\
AEScorer     & Qwen2.5-7B-Inst     & $59.75$                & $38.36$               \\ \hline
G-Verifier      & Qwen3-4B-Inst       & $41.79$                & $9.45$                \\
AG-Verifier     & Qwen3-4B-Inst       & $60.51$                & $41.40$               \\
AEScorer     & Qwen3-4B-Inst       & \textbf{65.41}       & \textbf{47.59}      \\ \hline
\end{tabular}}
\caption{Results on long-form factuality verification.}
\label{tab:document}
\end{table}

\subsection{Adaptability to Binary Verification}
While our focus is graded factuality verification, we also investigate AEScorer's effectiveness in the binary verification setting required in conventional scenarios.
We reconstructed GradedVeriBench with binary annotations (correct or incorrect)\footnote{Given that questions in GradedVeriBench include ground-truth answers, classifying answers as correct or incorrect was straightforward for human annotators.}. We then evaluated whether different verifiers could assign higher scores to correct examples than to incorrect ones. As shown in Table \ref{tab:binary}, AEScorer achieves superior performance, demonstrating that evidence-grounded scoring also transfers well to binary factuality verification.

\subsection{Application to Long-Form Verification}

In this section, we apply AEScorer to long-form texts, another scenario where factuality verification is desired. 
For this, we use the dataset from FactScore \cite{min2023factscore} which contains three long-form, model-generated responses for each question, annotated with factuality scores. Following the FactScore protocol, we first decomposed each response into atomic claims and then verified each claim individually. As shown in Table \ref{tab:document}, AEScorer also achieves the best long-form verification performance, outperforming FactScore despite using substantially smaller LLMs than GPT-4. This suggests that evidence-grounded graded scoring remains effective even when verification is applied after claim decomposition.

\subsection{The Influence of Knowledge Source}
As detailed in Section \ref{sec:architecture}, our agentic search incorporates two distinct knowledge sources: \texttt{WebSearch} and \texttt{LocalSearch}. This design enables the agent to adaptively retrieve information from the most appropriate source. To validate this, we conducted ablation studies on these sources.

As shown in Table \ref{tab:knowledge}, \texttt{WebSearch} significantly outperforms \texttt{LocalSearch}. This result aligns with intuition, as search engines provide broad, diverse, and real-time web content while leveraging advanced algorithms for precise retrieval. 
In contrast, \texttt{LocalSearch} relies on simple similarity matching over static and limited documents. Nonetheless, combining both sources yields the best performance, demonstrating the effectiveness of diverse knowledge sources for agentic search.

\begin{table}[t]
\centering
\resizebox{0.45 \textwidth}{!}{
\begin{tabular}{cccc}
\hline
\textbf{Method} & \textbf{Model}  & \textbf{Knowledge} & \textbf{Average} \\ \hline
D-Verifier      & Qwen2.5-7B-Inst & -                  & $51.87$            \\
AEScorer     & Qwen2.5-7B-Inst & WebSearch          & $79.46$            \\
AEScorer     & Qwen2.5-7B-Inst & LocalSearch        & $74.32$            \\
AEScorer     & Qwen2.5-7B-Inst & both               & \textbf{80.08}   \\ \hline
D-Verifier      & Qwen3-4B-Inst   & -                  & $46.92$            \\
AEScorer     & Qwen3-4B-Inst   & WebSearch          & $77.05$            \\
AEScorer     & Qwen3-4B-Inst   & LocalSearch        & $70.77$            \\
AEScorer     & Qwen3-4B-Inst   & both               & \textbf{78.69}   \\ \hline
\end{tabular}}
\caption{Kendall-tau correlation coefficient of different knowledge sources on GradedVeriBench.}
\label{tab:knowledge}
\end{table}

\begin{figure}[t]
    \centering
        \includegraphics[width=1.0\linewidth]{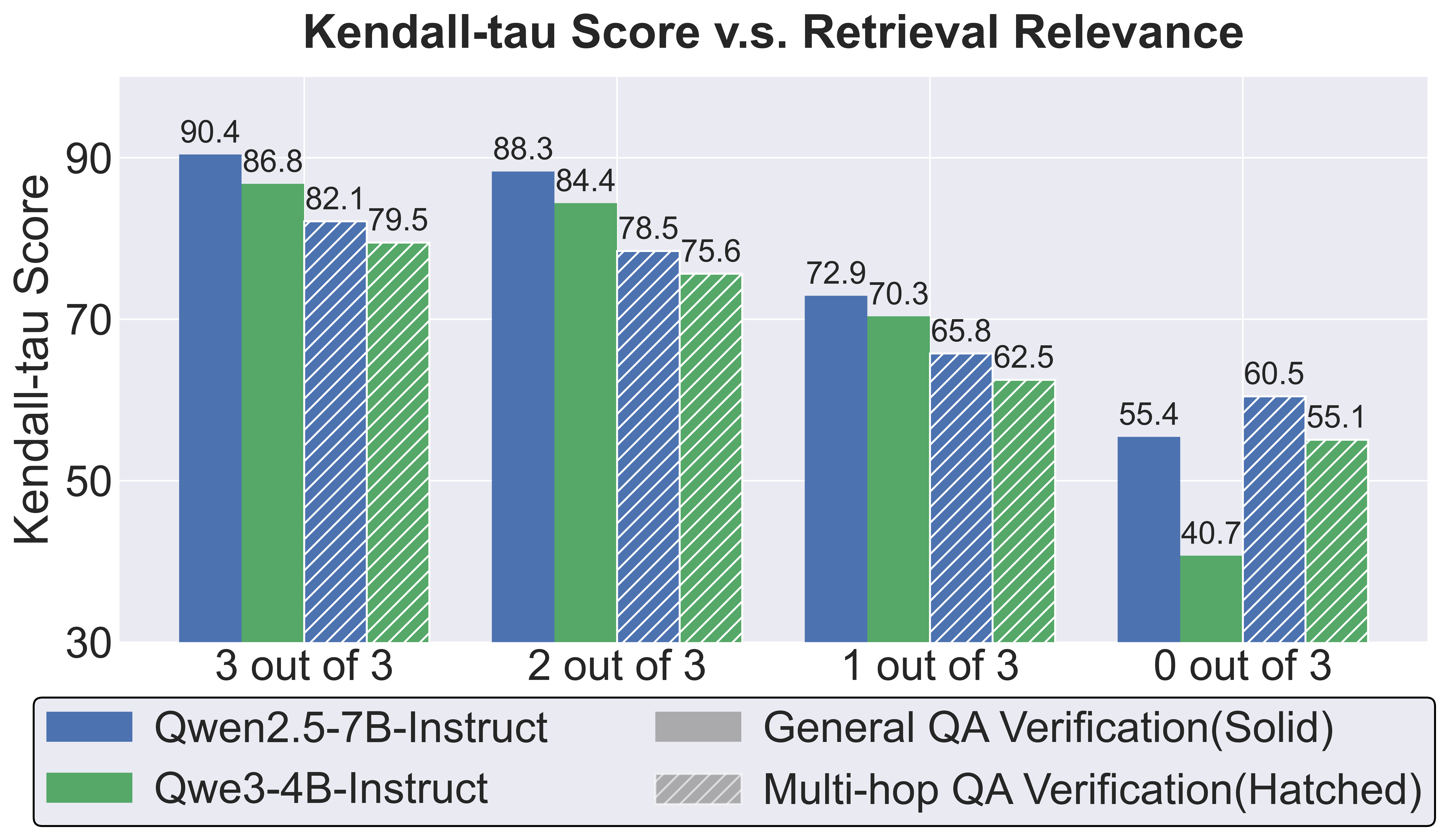}
        \caption{Impact of retrieval relevance on verification performance on GradedVeriBench. The x-axis represents the relevance level, defined as the number of candidate answers (0 to 3) supported by the retrieved context. The results demonstrate a clear trend that higher retrieval relevance leads to more accurate verification.}
    \label{figure:variation-knowledge}
\end{figure}

To further isolate the impact of external knowledge, we use GPT-4 to assess the relevance of the retrieved results for each sample. We then categorized samples based on their retrieval relevance levels\footnote{The relevance level is defined by the number of candidate answers to which the retrieved results are regarded as relevant.}, as shown in Figure \ref{figure:variation-knowledge}. The results reveal a strong correlation between retrieval relevance and final verification accuracy. This confirms the critical role of external knowledge: when retrieved information is irrelevant, verification accuracy declines sharply due to the lack of reliable evidence.
\section{Conclusion}

Graded factuality scoring is important because it provides denser factuality feedback for answers with various degrees of correctness. We propose AEScorer for graded factuality verification. AEScorer combines agentic evidence acquisition with graded scoring, enabling grounded verification for short answers. We also develop GradedVeriBench, a benchmark dedicated to this setting. Experimental results suggest that factuality evaluation can move beyond coarse correctness labels toward feedback signals that better reflect the graded nature of real-world factual errors. 

\section*{Limitation}
Our work currently has a few limitations that outline directions for future research:

\vspace{-2mm}
\begin{itemize}[leftmargin=4mm, itemsep=1mm, parsep=0pt]
    \item Agentic Capabilities: While we leveraged the inherent agentic search capabilities of Qwen-2.5-7B and Qwen-3-8B, we have not yet specifically optimized this behavior for factuality verification. Future work will focus on further enhancing and fine-tuning these search abilities.    
    \item Model Architecture: Although LoRA reduces parameter overhead, the current implementation of AEScorer still leaves room for tighter integration between evidence acquisition and scoring. Future work may further simplify the pipeline to reduce computational cost and latency.
    \item Application in Post-training: Although graded verification is promising for reward modeling and reranking, its downstream value in full post-training pipelines still requires broader validation across more models and tasks.
\end{itemize}

\section*{Ethics Statement}
This work uses publicly available QA benchmarks and focuses on evaluating factuality verification systems rather than generating new factual claims. To build the benchmark, we relied on a combination of LLM-based judging and human verification. The human annotation effort was conducted by graduate-student annotators following explicit guidelines, and such annotation should be fairly compensated under local research standards.

Our framework also inherits potential biases from its evidence sources and underlying models. Web search results, Wikipedia-based local corpora, and judge models may reflect historical, cultural, or distributional biases, which can affect factuality assessments. Therefore, AEScorer should be viewed as a decision-support tool rather than an unquestionable oracle, especially in high-stakes scenarios. We do not recommend using it as the sole basis for unsupervised decision making in domains such as medicine, law, or public policy.

\section*{Acknowledgments}
This work was supported by National Natural Science Foundation of China (62276077) and JST K Program Grant Number JPMJKP$24$C$3$, Japan. We also thank Lvyuan Han for his assistance in implementing the agentic retrieval system.

\bibliography{main}
\clearpage
\appendix

\definecolor{thoughtgray}{gray}{0.95}

\section{Implementation Details}
\label{app:implementations}
As mentioned in Section \ref{sec:disc-training}, we train the scoring capability of AEScorer with LoRA to reduce storage consumption. The training was conducted on 8 NVIDIA A6000-48GB GPUs with Transformers \cite{wolf2019huggingface} and DeepSpeed \cite{rasley2020deepspeed}. Detailed hyperparameters are presented in Table \ref{tab:hyperparams}. To adapt the base model into a factuality scorer, the prediction head is newly initialized as a linear projection layer\footnote{Please refer to \texttt{AutoModelForSequence}\texttt{Classification} in Huggingface library for more details.}.

\begin{table}[h]
\centering
\resizebox{0.48\textwidth}{!}{
\begin{tabular}{l|cc}
\hline
\textbf{Configuration} & \textbf{Qwen2.5-7B-Inst} & \textbf{Qwen3-4B-Inst} \\ \hline
ranking margin         & 0.1                      & 0.1                     \\
max length             & 1024                     & 1024                    \\
learning rate          & 2e-4                     & 1e-4                    \\
scheduler              & cosine decay             & cosine decay            \\
optimizer              & AdamW                    & AdamW                   \\
training epochs        & 3                        & 3                       \\
lora rank              & 8                        & 8                       \\
lora alpha             & 32                       & 32                      \\
lora dropout           & 0.05                     & 0.05                    \\
data types             & bf16                     & bf16                    \\
ZeRO optimizer         & stage 2                  & stage 2                 \\ \hline
\end{tabular}}
\caption{Settings of scoring optimization.}
\vspace{-3mm}
\label{tab:hyperparams}
\end{table}

For retrieval on local documents, we use the 2018 Wikipedia dump \cite{karpukhin2020dense} as the knowledge source and E5 \cite{wang2022text} as the retriever. We follow \citet{lin2022truthfulqameasuringmodelsmimic} and set the number of retrieved passages to 3 across all retrieval-based methods. For web retrieval, we use the Google Search API provided by Serper\footnote{\url{https://serper.dev}}, and we keep the top 10 snippets from the search results.

\section{Scaling the Scoring Component}
\label{app:scorer-scale}
This section examines whether the scoring component requires a large model once verification-oriented evidence has already been acquired. We vary the model size used for the scoring component in AEScorer while fixing the evidence acquisition backbone at $7$B, and compare this trend with verifiers of different model sizes.

\begin{figure}[t]
    \centering
        \includegraphics[width=1.0\linewidth]{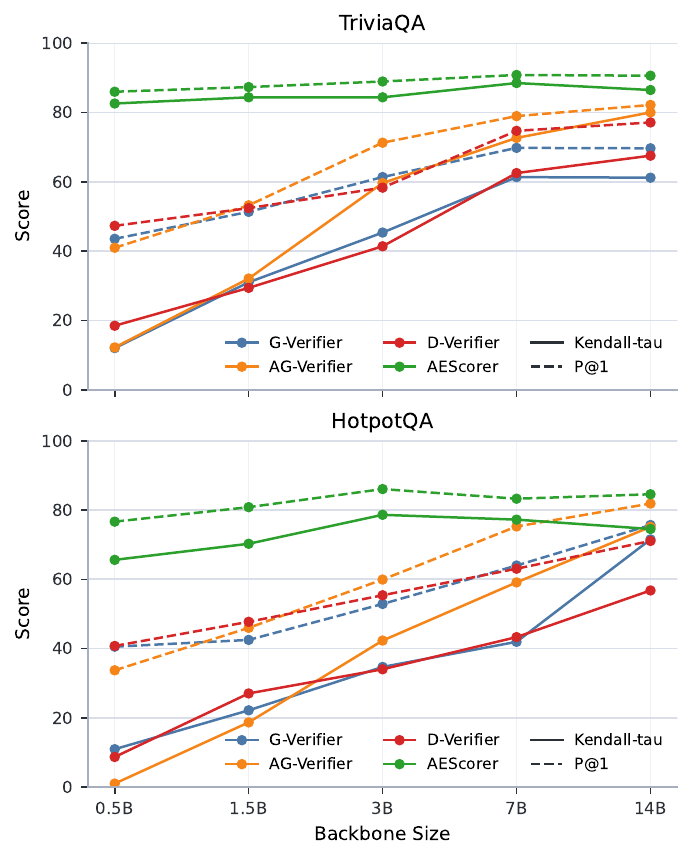}
        \caption{Variation of Kendall-tau and Precision@1 scores with different verifier sizes on TriviaQA (top) and HotpotQA (bottom).}
        \vspace{-3mm}
    \label{figure:variation-model}
\end{figure}

As shown in Figure \ref{figure:variation-model}, the generative verifier's performance consistently improves with scorer size, as the same model must conduct reasoning, tool use, and verdict generation. The naive discriminative verifier's performance also improves with model size, as it relies solely on internal knowledge for scoring, which scales with model capacity. 
In contrast, AEScorer remains relatively stable across different scorer sizes. Notably, AEScorer with a smaller scoring component still outperforms the largest AG-Verifier ($14$B), suggesting that explicit evidence grounding can reduce the scale requirement of the final scorer.

Moreover, the benefit of retrieval for generative verification also becomes larger as model size increases. Smaller models often lack sufficient planning and tool-use capabilities to retrieve useful evidence, whereas larger models can better exploit retrieved contexts during verdict generation.


\section{Scaling the Acquisition Component}
\label{app:retriever-scale}

\begin{figure}[t]
    \centering
        \includegraphics[width=1.0\linewidth]{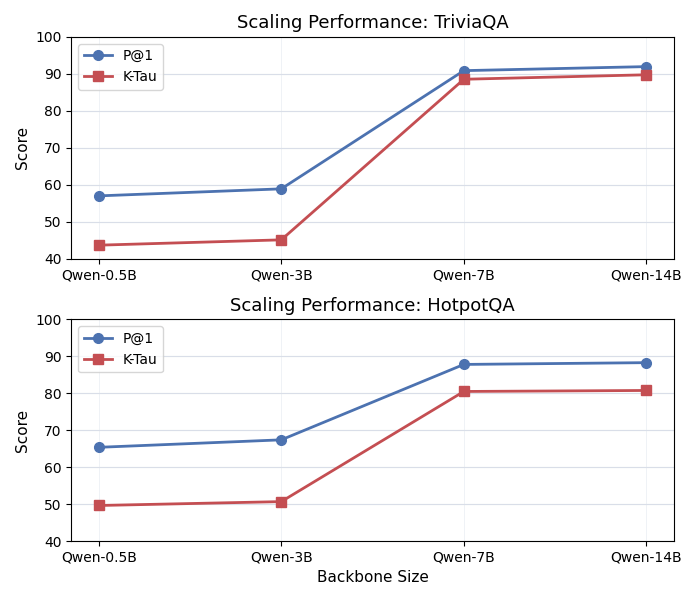}
        \caption{The influence of scaling the evidence acquisition backbone on verification accuracy.}
        \vspace{-3mm}
    \label{fig:acquisition-scale-metric}
\end{figure}

\begin{figure}[h]
    \centering
        \includegraphics[width=1.0\linewidth]{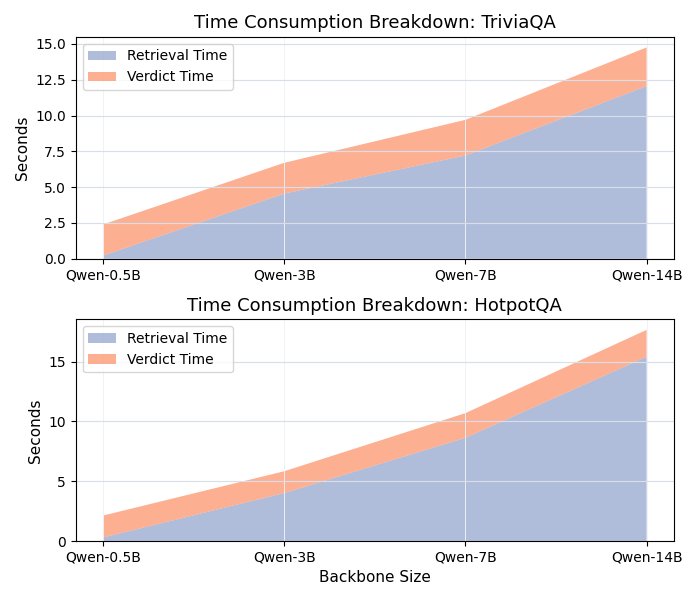}
        \caption{The influence of scaling the evidence acquisition backbone on time consumption.}
        \vspace{-3mm}
    \label{fig:acquisition-scale-time}
\end{figure}

\begin{figure}[h]
    \centering
        \includegraphics[width=1.0\linewidth]{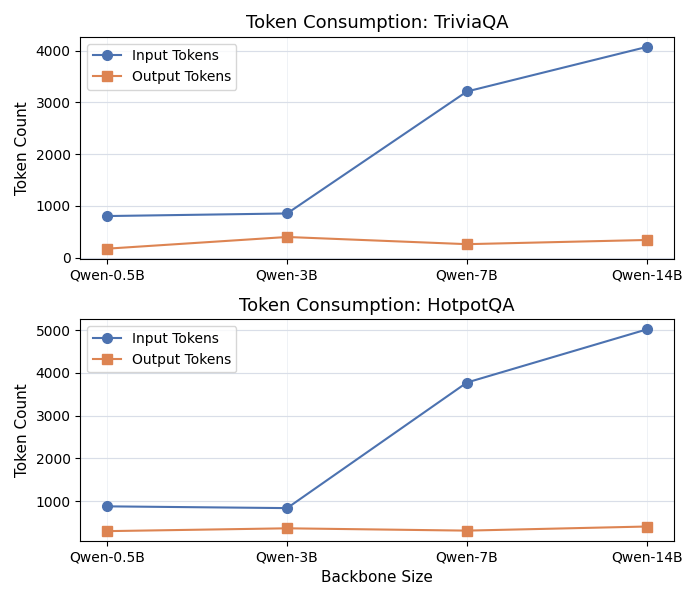}
        \caption{The influence of scaling the evidence acquisition backbone on token consumption.}
        \vspace{-3mm}
    \label{fig:acquisition-scale-token}
\end{figure}

Complementary to Appendix \ref{app:scorer-scale}, this section analyzes the effect of scaling the evidence acquisition backbone. We keep the scorer fixed and vary only the model used for agentic evidence acquisition, so that the experiment isolates how acquisition quality affects graded factuality scoring.

As shown in Figures \ref{fig:acquisition-scale-metric}, \ref{fig:acquisition-scale-token} and \ref{fig:acquisition-scale-time}, verification accuracy drops sharply when the acquisition backbone is too small, such as $0.5$B. This degradation occurs because small models often lack the planning, tool-use, and reasoning capabilities required for effective agentic search. As a result, they may retrieve incomplete evidence, fail to connect multi-hop facts, or introduce noisy trajectories that weaken the downstream scoring step.

Moreover, although performance generally improves with a stronger acquisition model, larger backbones also increase latency and token consumption. The gains from $7$B to $14$B are marginal, suggesting a ceiling effect: once the model can reliably collect the necessary external evidence, further scaling is limited by evidence availability rather than model capacity. Therefore, the $7$B acquisition backbone offers a favorable balance between verification accuracy and operational efficiency.

These results support a practical design principle for AEScorer: model capacity is more critical for evidence acquisition than for the final scoring step. An alternative implementation could therefore use a stronger model for acquisition and a lighter model for scoring, akin to the design of \citet{tang2024minicheck}. We leave this direction to future work.

\section{Efficiency Analysis}
\label{app:efficiency}

To better understand the computational trade-offs of agentic evidence acquisition, we compare AEScorer against the Discriminative Verifier (D-Verifier), Generative Verifier (G-Verifier), and Agentic-Generative Verifier (AG-Verifier). We report token usage, runtime, retrieval iterations, and verification accuracy for a single response.

As shown in Table \ref{tab:efficiency}, D-Verifier is the fastest, as it involves neither generation nor retrieval processes. The G-Verifier also maintains a relatively high speed because it requires only a single generation pass per input without retrieval. However, both models suffer from poor verification accuracy due to their lack of external knowledge.

Compared with AG-Verifier, AEScorer achieves higher verification accuracy with lower token overhead and shorter runtime. Since the two methods share the same acquisition process, this gap mainly comes from the final verdict stage: AEScorer replaces free-form verdict generation with a directly optimized scoring step. This makes the final decision both faster and more precise for graded factuality verification.

\section{Case Study}
\label{app:case-study}

In this section, we conduct a qualitative case study to compare the G-Verifier with AEScorer. We selected two samples from the Natural Questions \cite{kwiatkowski2019natural} and Musique datasets \cite{trivedi-etal-2022-musique}, representing general QA and multi-hop QA verification respectively.

\begin{table*}[t]
\centering
\resizebox{0.98\textwidth}{!}{
\begin{tabular}{llcccccccc}
\hline
\textbf{Dataset} & \textbf{Verifier} & \textbf{Tokens-Input} & \textbf{Tokens-Output} & \textbf{Time-Total} & \textbf{Time-Retrieval} & \textbf{Time-Verdict} & \textbf{Retrieval-Iter} & \textbf{P@1} & \textbf{K-Tau} \\ \hline
\multirow{4}{*}{TriviaQA} 
& D-Verifier & 472 & 0 & 0.02 & 0 & 0.02 & 0 & 75.81 & 63.80 \\
& G-Verifier & 472 & 306 & 5.66 & 0 & 5.66 & 0 & 69.81 & 61.37 \\
& AG-Verifier & 8676 & 527 & 15.82 & 7.21 & 8.61 & 3.02 & 78.98 & 72.69 \\
& AEScorer & 4884 & 365 & 9.70 & 7.21 & 2.49 & 3.02 & \textbf{90.86} & \textbf{88.53} \\ \hline
\multirow{4}{*}{HotpotQA} 
& D-Verifier & 478 & 0 & 0.02 & 0 & 0.02 & 0 & 63.76 & 45.18 \\
& G-Verifier & 478 & 335 & 6.71 & 0 & 6.71 & 0 & 63.99 & 41.96 \\
& AG-Verifier & 9773 & 586 & 16.98 & 8.63 & 8.35 & 3.56 & 75.26 & 59.12 \\
& AEScorer & 5611 & 420 & 10.69 & 8.63 & 2.06 & 3.56 & \textbf{87.80} & \textbf{80.49} \\ \hline
\end{tabular}}
\caption{Efficiency comparison of different verifiers. AEScorer retains the same retrieval cost as AG-Verifier but substantially reduces the cost of the final verdict stage.}
\label{tab:efficiency}
\end{table*}

As illustrated in Figure \ref{fig:naive_single} and \ref{fig:naive_multi}, the Generative Verifier relies solely on parametric memory, a limitation that frequently leads to factual hallucinations. For instance, in Figure \ref{fig:naive_single}, the model fails to adhere to strict terminological constraints, incorrectly favoring the Greek god "Hades" over the correct Roman equivalent "Dis Pater." Moreover, in Figure \ref{fig:naive_multi}, the model suffers from knowledge conflation, hallucinating "Ponte Vecchio" (a bridge in Florence) as the answer for a query about Venice. These failures demonstrate the unreliability of relying solely on internal model knowledge for fact-checking.

Conversely, as shown in Figure \ref{fig:agentic_single} and \ref{fig:agentic_multi}, AEScorer leverages reasoning and tool use to acquire necessary knowledge from both local documents and web search engines. This mechanism compensates for internal knowledge deficits and provides grounded context for graded scoring. Moreover, AEScorer is particularly advantageous for multi-hop QA because it can perform targeted decomposed searches. Rather than solving the complex query in a single pass, the agent first identifies the composer, then the birthplace, and finally the bridge, thereby grounding the final verification in retrieved facts.

\section{Failure Analysis}
\label{app:failure-analysis}

Although agentic search generally improves verification, AEScorer remains sensitive to the relevance and correctness of retrieved evidence. To demonstrate this, We manually inspect the errors made by the Qwen2.5-7B-Instruct-based AEScorer on HotpotQA and identify a representative failure caused by partially related but misleading retrieval results.

\begin{quote}
\small
\textbf{Question:} What year did the father of Willem van Oldenbarnevelt die?\\
\textbf{Answer to verify:} The father of Willem van Oldenbarnevelt died in 1619.\\[2pt]
\textbf{Search trace summary:} The model first searched for ``father of Willem van Oldenbarnevelt.'' The retrieved local document discussed Reinier van Oldenbarnevelt and an assassination attempt related to his father. It then searched for ``Reinier van Oldenbarnevelt death year,'' retrieving results that mentioned 1623 and 1 October 1636.\\[2pt]
\textbf{Compressed verdict:} The model treated Reinier as Willem's father, concluded that the evidence did not support 1619, and predicted \textit{Incorrect}.
\end{quote}

The retrieved information concerns Reinier van Oldenbarnevelt, who was Willem's brother rather than his father. Because the evidence was lexically related but referred to the wrong family relation, the refinement stage retained incorrect useful facts, which then propagated to the final score. This case illustrates that retrieval does not guarantee reliable verification. Instead, partially relevant but misleading evidence can be especially harmful when the scorer lacks sufficient prior knowledge.

\section{Data Sample of GradedVeriBench}
\label{app:data-sample}
In this section, we present two representative samples from GradedVeriBench. As illustrated in Figures \ref{fig:data-sample1} and \ref{fig:data-sample2}, the dataset exhibits a strict factuality hierarchy: Rank 1 represents the highest level of factual accuracy, Rank 2 exhibits moderate factuality, and Rank 3 incorporates severe factual errors. This design supports robust evaluation of graded factuality verification.

\section{Prompt templates}
\label{sec:app-prompt}
In this section, we present the detailed prompt templates for both the generative baselines and our proposed AEScorer. The prompts for Step 1 and Step 2 of AEScorer are presented in Prompt \ref{prompt:agentic_search} and \ref{prompt:context_compression}, while the prompts for the Generative Verifier and Generative Scorer are presented in Prompt \ref{prompt:naive_generative_verification}, \ref{prompt:generative_verification}, and \ref{prompt:generative_scoring}.

\begin{figure*}
    \centering
    \scriptsize
    \fontfamily{cmr}\selectfont 
    \begin{tcolorbox}[
        enhanced,
        colback=white,
        colframe=gray!60!black,
        title=\textbf{\normalsize Naive Generative Verifier: Full Trajectory},
        fonttitle=\bfseries,
        boxrule=1pt
    ]

    \begin{tcolorbox}[
        colback=gray!5, 
        colframe=gray!30, 
        left=2pt, right=2pt, top=2pt, bottom=2pt
    ]
        \textbf{Question:} roman god of the underworld also called orcus or pluto? \\
        \textbf{Candidates:} \textbf{Answer 1:} Pluto \quad \textbf{Answer 2:} Hades \quad \textbf{Answer 3:} Roman god... is Dis Pater. \\
        \textbf{Annotation:} \textbf{Answer 3} $>$ \textbf{Answer 1} $>$ \textbf{Answer 2}
    \end{tcolorbox}
    
    \vspace{0.15cm}
    \hrule
    \vspace{0.15cm}

    \textbf{\textsc{Step 1: Direct Verify}}
    \vspace{2pt}

    \colorbox{thoughtgray}{\parbox{\dimexpr\linewidth-2\fboxsep}{
        \textbf{Explanation:}
        \begin{itemize}[leftmargin=10pt, noitemsep, topsep=0pt]
            \item \textbf{Answer 1:} While Pluto is correct in the context of Roman mythology, it is less precise than Hades because it does not specify the Greek origin of the god. However, it is still a correct answer.
            \item \textbf{Answer 2:} This is the most accurate answer as Hades is the Greek god of the underworld, and he is the primary deity associated with the underworld in Greek mythology. In Roman mythology, Hades is known as Pluto.
            \item \textbf{Answer 3:} Roman god of the underworld also called Orcus or Pluto is Dis Pater**: This answer is also correct but includes additional information (Dis Pater) that is not directly relevant to the question. It is less concise and could be misleading if the focus is solely on the Roman god of the underworld.
        \end{itemize}
        Based on this, Answer 2 is the most factually correct, followed by Answer 1, and Answer 3 is incorrect.
    }}
    \vspace{1pt}
    
    \begin{tcolorbox}[
        colback=scorelow!40, 
        colframe=gray!30, 
        left=2pt, right=2pt, top=4pt, bottom=4pt
    ]
        \textbf{Final Verdict:} Answer 3 $>$ Answer 1 $>$ Answer 2
    \end{tcolorbox}
    
    \end{tcolorbox}
    \caption{Full execution trace of the naive generative verification pipeline for general QA.}
    \label{fig:naive_single}
\end{figure*}

\begin{figure*}
    \centering
    \scriptsize 
    \fontfamily{cmr}\selectfont 
    \begin{tcolorbox}[
        enhanced,
        colback=white,
        colframe=gray!60!black,
        title=\textbf{\normalsize Naive Generative Verifier: Full Trajectory},
        fonttitle=\bfseries,
        boxrule=1pt
    ]


    \begin{tcolorbox}[
        colback=gray!5, 
        colframe=gray!30, 
        left=2pt, right=2pt, top=2pt, bottom=2pt
    ]
        \textbf{Question:} What is the name of the famous bridge in the city where the composer of ``La costanza trionfante degl'amori e de gl'odii'' was born? \\
        \textbf{Candidates:} \textbf{Answer 1:} Ponte di Rialto \quad \textbf{Answer 2:} Venice \quad \textbf{Answer 3:} The name... is the Ponte Vecchio. \\
        \textbf{Annotation:} \textbf{Answer 1} $>$ \textbf{Answer 2} $>$ \textbf{Answer 3}
    \end{tcolorbox}
    
    \vspace{0.15cm}
    \hrule
    \vspace{0.15cm}

    \textbf{\textsc{Step 1: Direct Verify}}
    \vspace{2pt}

    \colorbox{thoughtgray}{\parbox{\dimexpr\linewidth-2\fboxsep}{
        \textbf{Explanation:}
        \begin{itemize}[leftmargin=10pt, noitemsep, topsep=0pt]
            \item \textbf{Answer 3:} ``The name... is the Ponte Vecchio.'' This answer is factually correct. Antonio Vivaldi... was born in Venice, and the Ponte Vecchio is indeed a famous bridge in Venice.
            \item \textbf{Answer 1:} ``Ponte di Rialto'' - This answer is also factually correct, as the Ponte di Rialto is a famous bridge in Venice. However, it is less specific than Answer3, which directly states the name of the bridge.
            \item \textbf{Answer 2:} ``Venice'' - This answer is not specific enough. While Venice is the city where the composer was born, it does not specify the name of the famous bridge. Therefore, it is less factually correct than Answer3.
        \end{itemize}
        Therefore, Answer3 is the most factually correct, followed by Answer1, and Answer2 is the least factually correct.
    }}
    \vspace{1pt}
    
    \begin{tcolorbox}[
        colback=scorelow!40, 
        colframe=gray!30, 
        left=2pt, right=2pt, top=4pt, bottom=4pt
    ]
        \textbf{Final Verdict:} Answer 3 $>$ Answer 1 $>$ Answer 2
    \end{tcolorbox}
    
    \end{tcolorbox}
    \caption{Full execution trace of the naive generative verification pipeline for multi-hop QA.}
    \label{fig:naive_multi}
\end{figure*}

\begin{figure*}
    \centering
    \scriptsize
    \fontfamily{cmr}\selectfont 
    \begin{tcolorbox}[
        enhanced,
        colback=white,
        colframe=gray!60!black,
        title=\textbf{\normalsize AEScorer: Full Trajectory},
        fonttitle=\bfseries,
        boxrule=1pt
    ]

    \begin{tcolorbox}[
        colback=gray!5, 
        colframe=gray!30, 
        left=2pt, right=2pt, top=2pt, bottom=2pt
    ]
        \textbf{Question:} roman god of the underworld also called orcus or pluto? \\
        \textbf{Candidates:} \textbf{Answer 1:} Pluto \quad \textbf{Answer 2:} Hades \quad \textbf{Answer 3:} Roman god... is Dis Pater. \\
        \textbf{Annotation:} \textbf{Answer 3} $>$ \textbf{Answer 1} $>$ \textbf{Answer 2}
    \end{tcolorbox}
    
    \vspace{0.15cm}
    \hrule
    \vspace{0.15cm}

    \textbf{\textsc{Step 1: Agentic Search}}
    \vspace{2pt}
    
    \colorbox{thoughtgray}{\parbox{\dimexpr\linewidth-2\fboxsep}{%
        \textbf{Thought:} I need to identify the Roman god of the underworld who is synonymous with Orcus or Pluto. I will start by searching local knowledge for these specific names.
    }}
    \vspace{1pt}
    \colorbox{toolblue}{\parbox{\dimexpr\linewidth-2\fboxsep}{%
        \textbf{Tool Call:} \texttt{search\_local(``roman god of the underworld orcus pluto'')}
    }}
    \vspace{1pt}
    \colorbox{evidcream}{\parbox{\dimexpr\linewidth-2\fboxsep}{%
        \textbf{Observations:} \textit{[Source: Local Wikipedia]}
        \begin{itemize}[leftmargin=10pt, noitemsep, topsep=0pt]
            \item \textbf{Doc 2 (``Orcus''):} ...Orcus was a god of the underworld... In later tradition, he was conflated with Dis Pater.
            \item \textbf{Doc 3 (``Dīs Pater''):} ...Dīs Pater was a Roman god of the underworld... later equated with the chthonic deities Pluto (Hades) and Orcus.
        \end{itemize}
    }}

    \vspace{3pt}
    \colorbox{thoughtgray}{\parbox{\dimexpr\linewidth-2\fboxsep}{%
        \textbf{Thought:} Local results suggest Dis Pater, Pluto, and Orcus are conflated/synonymous. I will verify this with a web search to ensure these names refer to the same entity in this context.
    }}
    \vspace{1pt}
    \colorbox{toolblue}{\parbox{\dimexpr\linewidth-2\fboxsep}{%
        \textbf{Tool Call:} \texttt{search\_web(``roman god of the underworld orcus pluto'')}
    }}
    \vspace{1pt}
    \colorbox{evidcream}{\parbox{\dimexpr\linewidth-2\fboxsep}{%
        \textbf{Observations:} \textit{[Source: Google Search]}
        \begin{itemize}[leftmargin=10pt, noitemsep, topsep=0pt]
            \item \textbf{Result 2:} In ancient Roman religion... Pluto also known as \textbf{Dis Pater} or Orcus, was the god of the dead...
            \item \textbf{Result 4:} \textbf{Dis Pater}, also known as Orcus, is the Roman god of the infernal regions, equivalent to the Greek Hades or Pluto.
        \end{itemize}
    }}

    \vspace{3pt}
    \colorbox{thoughtgray}{\parbox{\dimexpr\linewidth-2\fboxsep}{%
        \textbf{Thought:} The search results consistently confirm that Pluto, Orcus, and Dis Pater are synonymous names for the Roman god of the underworld. Hades is the Greek equivalent. \textbf{READY\_FOR\_ANSWERING}
    }}

    \vspace{0.15cm}
    \hrule
    \vspace{0.15cm}

    \textbf{\textsc{Step 2: Context Compression}}
    \vspace{2pt}
    
    \begin{tcolorbox}[colback=white, colframe=gray!30, left=1pt, right=1pt, top=1pt, bottom=1pt]
        \textbf{Compressed Context for Answer 1:} \\
        \textbf{Useful Facts:} Pluto is the Roman god of the underworld; Pluto is known as Dis Pater or Orcus; The names are interchangeable, but Dis Pater is often the formal Roman title. \\
        \textbf{Reasoning:} 1. Search results consistently mention Pluto as the Roman god. 2. Information indicates Orcus and Dis Pater are conflated with Pluto. 3. Multiple sources confirm they are synonymous referring to the same god.
    \end{tcolorbox}
    \vspace{-3pt}
    
    \begin{tcolorbox}[colback=white, colframe=gray!30, left=1pt, right=1pt, top=1pt, bottom=1pt]
        \textbf{Compressed Context for Answer 2:} \\
        \textbf{Useful Facts:} Hades is the Greek god of the underworld; The Roman equivalent is Pluto/Dis Pater; Hades is not the correct name in a strictly Roman context. \\
        \textbf{Reasoning:} 1. Orcus and Pluto are Roman gods. 2. Dis Pater is synonymous with them. 3. Hades is identified as the Greek god, not Roman, making this answer incorrect for the specific query.
    \end{tcolorbox}

    \begin{tcolorbox}[colback=white, colframe=gray!30, left=1pt, right=1pt, top=1pt, bottom=1pt]
        \textbf{Compressed Context for Answer 3:} \\
        \textbf{Useful Facts:} Dis Pater is explicitly stated to be the Roman god of the underworld; Dis Pater is synonymous with Orcus and Pluto; All three names refer to the same deity in Roman mythology. \\
        \textbf{Reasoning:} 1. Search results confirm Orcus, Pluto, and Dis Pater are all Roman gods of the underworld. 2. Sources indicate Orcus and Pluto are synonymous. 3. Dis Pater is explicitly stated as equivalent to Pluto and Hades.
    \end{tcolorbox}
    \vspace{-3pt}

    \vspace{0.1cm}
    \hrule
    \vspace{0.1cm}

    \textbf{\textsc{Step 3: Score Prediction}}
    \vspace{2pt}
    
    \begin{center}
    \begin{tabular}{p{0.3\textwidth} p{0.3\textwidth} p{0.3\textwidth}}
        \colorbox{scoremid}{\textbf{Answer 1: 0.4580}} & 
        \colorbox{scorelow}{\textbf{Answer 2: 0.2188}} &
        \colorbox{scorehigh}{\textbf{Answer 3: 0.5409}} \\
    \end{tabular}
    \end{center}

    \end{tcolorbox}
    \caption{Full execution trace of agentic discriminative verification for general QA.}
    \label{fig:agentic_single}
\end{figure*}

\begin{figure*}[h!]
    \centering
    \scriptsize 
    \fontfamily{cmr}\selectfont 
    \begin{tcolorbox}[
        enhanced,
        colback=white,
        colframe=gray!60!black,
        title=\textbf{\normalsize AEScorer: Full Trajectory},
        fonttitle=\bfseries,
        boxrule=1pt
    ]


    \begin{tcolorbox}[
        colback=gray!5, 
        colframe=gray!30, 
        left=2pt, right=2pt, top=2pt, bottom=2pt
    ]
        \textbf{Question:} What is the name of the famous bridge in the city where the composer of ``La costanza trionfante degl'amori e de gl'odii'' was born? \\
        \textbf{Candidates:} \textbf{Answer 1:} Ponte di Rialto \quad \textbf{Answer 2:} Venice \quad \textbf{Answer 3:} The name... is the Ponte Vecchio. \\
        \textbf{Annotation:} \textbf{Answer 1} $>$ \textbf{Answer 2} $>$ \textbf{Answer 3}
    \end{tcolorbox}
    
    \vspace{0.15cm}
    \hrule
    \vspace{0.15cm}

    \textbf{\textsc{Step 1: Agentic Search}}
    \vspace{2pt}
    
    \colorbox{thoughtgray}{\parbox{\dimexpr\linewidth-2\fboxsep}{%
        \textbf{Thought:} First, I need to find out in which city the composer of ``La costanza trionfante degl'amori e de gl'odii'' was born. Let's start by searching for information about this opera and its composer.
    }}
    \vspace{1pt}
    \colorbox{toolblue}{\parbox{\dimexpr\linewidth-2\fboxsep}{%
        \textbf{Tool Call:} \texttt{search\_local(``composer of La costanza trionfante degl'amori e de gl'odii'')}
    }}
    \vspace{1pt}
    \colorbox{evidcream}{\parbox{\dimexpr\linewidth-2\fboxsep}{%
        \textbf{Observations:} \textit{[Source: Local Wikipedia]}
        \begin{itemize}[leftmargin=10pt, noitemsep, topsep=0pt]
            \item \textbf{Doc 2 (``Griselda''):} ...Griselda is a dramma per musica... composed by Antonio Vivaldi.
            \item \textbf{Doc 3 (``Antonio Vivaldi''):} ...Also in 1716, Vivaldi wrote... ``La costanza trionfante...'' (RV 706).
        \end{itemize}
    }}

    \vspace{3pt}
    \colorbox{thoughtgray}{\parbox{\dimexpr\linewidth-2\fboxsep}{%
        \textbf{Thought:} The search result does not directly mention the composer... However, it mentions that Antonio Vivaldi... Let's search for more information about it to confirm if it is the same opera and find the composer's birthplace.
    }}
    \vspace{1pt}
    \colorbox{toolblue}{\parbox{\dimexpr\linewidth-2\fboxsep}{%
        \textbf{Tool Call:} \texttt{search\_web(``composer of La costanza trionfante degli amori e degli odi'')}
    }}
    \vspace{1pt}
    \colorbox{evidcream}{\parbox{\dimexpr\linewidth-2\fboxsep}{%
        \textbf{Observations:} \textit{[Source: Google Search]}
        \begin{itemize}[leftmargin=10pt, noitemsep, topsep=0pt]
            \item \textbf{Result 1:} Also in 1716, Vivaldi wrote and produced... La costanza trionfante... (RV 706).
            \item \textbf{Result 2:} La costanza trionfante... is a dramma per musica by \textbf{Antonio Vivaldi}.
        \end{itemize}
    }}

    \vspace{3pt}
    \colorbox{thoughtgray}{\parbox{\dimexpr\linewidth-2\fboxsep}{%
        \textbf{Thought:} The information confirms that Antonio Vivaldi was the composer of ``La costanza trionfante...''. Now, I need to find out where Antonio Vivaldi was born. Let's search for his birthplace.
    }}
    \vspace{1pt}
    \colorbox{toolblue}{\parbox{\dimexpr\linewidth-2\fboxsep}{%
        \textbf{Tool Call:} \texttt{search\_web(``birthplace of Antonio Vivaldi'')}
    }}
    \vspace{1pt}
    \colorbox{evidcream}{\parbox{\dimexpr\linewidth-2\fboxsep}{%
        \textbf{Observations:} \textit{[Source: Google Search]}
        \begin{itemize}[leftmargin=10pt, noitemsep, topsep=0pt]
            \item \textbf{Result 1:} Antonio Lucio Vivaldi was born on 4 March 1678 in \textbf{Venice}, then the capital...
            \item \textbf{Result 3:} The great composer... was indeed born in \textbf{Venice} and lived and worked here...
        \end{itemize}
    }}

    \vspace{3pt}
    \colorbox{thoughtgray}{\parbox{\dimexpr\linewidth-2\fboxsep}{%
        \textbf{Thought:} Based on the search results, Antonio Vivaldi was born in Venice, Italy. Now, I need to find the name of the famous bridge in Venice. Let's search for a famous bridge in Venice.
    }}
    \vspace{1pt}
    \colorbox{toolblue}{\parbox{\dimexpr\linewidth-2\fboxsep}{%
        \textbf{Tool Call:} \texttt{search\_web(``famous bridge in Venice'')}
    }}
    \vspace{1pt}
    \colorbox{evidcream}{\parbox{\dimexpr\linewidth-2\fboxsep}{%
        \textbf{Observations:} \textit{[Source: Google Search]}
        \begin{itemize}[leftmargin=10pt, noitemsep, topsep=0pt]
            \item \textbf{Result 1:} The \textbf{Rialto Bridge} (Ponte di Rialto)... is the oldest of the four bridges...
            \item \textbf{Result 8:} It is also the \textbf{most famous} in Venice thanks to its peculiar history...
        \end{itemize}
    }}
    
    \vspace{3pt}
    \colorbox{thoughtgray}{\parbox{\dimexpr\linewidth-2\fboxsep}{%
        \textbf{Thought:} Based on the search results, the Rialto Bridge is the most famous bridge in Venice. Therefore, the answer... is the Rialto Bridge. \textbf{READY\_FOR\_ANSWERING}
    }}

    \vspace{0.15cm}
    \hrule
    \vspace{0.15cm}

    \textbf{\textsc{Step 2: Context Compression}}
    \vspace{2pt}
    
    \begin{tcolorbox}[colback=white, colframe=gray!30, left=1pt, right=1pt, top=1pt, bottom=1pt]
        \textbf{Compressed Context for Answer 1:} \\
        \textbf{Useful Facts:} 
        Antonio Vivaldi was the composer of ``La costanza trionfante...''; 
        Antonio Vivaldi was born in Venice, Italy; 
        The Rialto Bridge (Ponte di Rialto) is the oldest and most famous bridge in Venice. \\
        \textbf{Reasoning:} 
        1. The search history confirms that Antonio Vivaldi was the composer of ``La costanza trionfante...''. 
        2. Further searches reveal that Antonio Vivaldi was born in Venice... 
        3. The search results also indicate that the Rialto Bridge... is the most famous bridge in Venice.
    \end{tcolorbox}
    \vspace{-3pt}
    
    \begin{tcolorbox}[colback=white, colframe=gray!30, left=1pt, right=1pt, top=1pt, bottom=1pt]
        \textbf{Compressed Context for Answer 2:} \\
        \textbf{Useful Facts:} 
        Antonio Vivaldi was the composer of ``La costanza trionfante...''; 
        Antonio Vivaldi was born in Venice...; 
        The Rialto Bridge is the most famous bridge in Venice. \\
        \textbf{Reasoning:} 
        1. The search history confirmed that Antonio Vivaldi was the composer of ``La costanza trionfante...''. 
        2. Further searches revealed that Antonio Vivaldi was born in Venice... 
        3. The search for famous bridges in Venice identified the Rialto Bridge as the most famous one.
    \end{tcolorbox}
    \vspace{-3pt}
    
    \begin{tcolorbox}[colback=white, colframe=gray!30, left=1pt, right=1pt, top=1pt, bottom=1pt]
        \textbf{Compressed Context for Answer 3:} \\
        \textbf{Useful Facts:} 
        Antonio Vivaldi was the composer of ``La costanza trionfante...''; 
        Antonio Vivaldi was born in Venice...; 
        The Rialto Bridge is the oldest and most famous bridge in Venice. \\
        \textbf{Reasoning:} 
        1. The search history confirms that Antonio Vivaldi was the composer of ``La costanza trionfante...''. 
        2. The search history also confirms that Antonio Vivaldi was born in Venice... 
        3. The search history provides information about famous bridges... specifically mentioning the Rialto Bridge...
    \end{tcolorbox}

    \vspace{0.1cm}
    \hrule
    \vspace{0.1cm}

    \textbf{\textsc{Step 3: Score Prediction}}
    \vspace{2pt}
    
    \begin{center}
    \begin{tabular}{p{0.3\textwidth} p{0.3\textwidth} p{0.3\textwidth}}
        \colorbox{scorehigh}{\textbf{Answer 1: 0.4448}} & 
        \colorbox{scoremid}{\textbf{Answer 2: 0.3479}} & 
        \colorbox{scorelow}{\textbf{Answer 3: 0.2119}} \\
    \end{tabular}
    \end{center}

    \end{tcolorbox}
    \caption{Full execution trace of agentic discriminative verification for multi-hop QA.}
    \label{fig:agentic_multi}
\end{figure*}

\begin{figure*}[t]
    \centering
    \scriptsize
    \fontfamily{cmr}\selectfont 
    \begin{tcolorbox}[
        enhanced,
        colback=white,
        colframe=gray!60!black,
        title=\textbf{\normalsize Data Sample 1},
        fonttitle=\bfseries,
        boxrule=1pt
    ]
        \textbf{Instruction:} What nationality is Alice Delamar's father? \\
        \textbf{Reference:} United States
        
        \vspace{0.2cm}
        \hrule
        \vspace{0.2cm}

        \colorbox{thoughtgray}{\parbox{\dimexpr\linewidth-2\fboxsep}{
            \begin{itemize}[leftmargin=35pt, noitemsep, topsep=2pt]
                \item[\textbf{Rank 1:}] American
                \item[\textbf{Rank 2:}] I don't have information on Alice Delamar.
                \item[\textbf{Rank 3:}] Alice Delamar's father's nationality is French.
            \end{itemize}
        }}
    \end{tcolorbox}
    \caption{A data sample constructed from 2Wiki \cite{xanh2020_2wikimultihop} in GradedVeriBench.}
    \vspace{-3mm}
    \label{fig:data-sample1}
\end{figure*}

\begin{figure*}[t]
    \centering
    \scriptsize
    \fontfamily{cmr}\selectfont 
    \begin{tcolorbox}[
        enhanced,
        colback=white,
        colframe=gray!60!black,
        title=\textbf{\normalsize Data Sample 2},
        fonttitle=\bfseries,
        boxrule=1pt
    ]
        \textbf{Instruction:} In the King's Speech, who played the person who was the King of England in 1950? \\
        \textbf{Reference:} Colin Firth
        
        \vspace{0.2cm}
        \hrule
        \vspace{0.2cm}

        \colorbox{thoughtgray}{\parbox{\dimexpr\linewidth-2\fboxsep}{
            \begin{itemize}[leftmargin=35pt, noitemsep, topsep=2pt]
                \item[\textbf{Rank 1:}] In the the King's Speech, the actor who played the person who was King of England in 1950 is Colin Firth.
                \item[\textbf{Rank 2:}] In the King's Speech, which is set in the 1930s, Colin Firth played the King of England, George VI. However, the question pertains to 1950, and by then, the King of England was still George VI, portrayed by Colin Firth in the film.
                \item[\textbf{Rank 3:}] George VI
            \end{itemize}
        }}
    \end{tcolorbox}
    \caption{A data sample constructed from Musique \cite{trivedi-etal-2022-musique} in GradedVeriBench.}
    \vspace{-3mm}
    \label{fig:data-sample2}
\end{figure*}

\begin{figure*}[]
\begin{promptbox}[title={Prompt ~\thetcbcounter: System Prompt for Agentic Search},
label={prompt:agentic_search}]
Your task is to determine the correct factual ranking of the provided answers. Based on the question and answers, think and identify what information you need and generate search queries using the available tools.\\

Question: \textbf{\{\{question\}\}}

\textbf{\{\{answers\_block\}\}}\\

Leverage both tools (`search\_local' for Wikipedia, `search\_web' for Google) and generate one search query per turn. If you have enough information, respond with `READY\_FOR\_EVALUATION'.
\end{promptbox}
\end{figure*}

\begin{figure*}
\begin{promptbox}[title={Prompt ~\thetcbcounter: System Prompt for Context Compression},
label={prompt:context_compression}]
You are a fact-checking expert. Below is a history of search actions performed by an agent to gather information about a specific question and answer. Your task is to analyze this history and determine the factual correctness of the answer.\\

\#\#\# SEARCH HISTORY \#\#\#

\textbf{\{\{search\_history\}\}}\\

\#\#\# VERIFICATION TASK \#\#\#

Based on the information from the search history, please verify the following:

**Question:** \textbf{\{\{question\}\}}

**Answer to Verify:** \textbf{\{\{answer\}\}}\\

\#\#\# ANALYSIS AND VERDICT \#\#\#

Based on your analysis, provide a structured response in the following format. Do not add any other text outside this structure.

**Useful Facts:** [List key facts from the search history relevant to the answer, separated by semicolons. Example: Fact1; Fact2; Fact3;]

**Reasoning:** [Provide a step-by-step reasoning for your verdict based on the useful facts.]

**Final Verdict:** [Your verdict: Correct, Incorrect, or Intermediate]
\end{promptbox}
\end{figure*}

\begin{figure*}[t]
\begin{promptbox}[title={Prompt ~\thetcbcounter: System Prompt for Naive Generative Verification},
label={prompt:naive_generative_verification}]
You are an expert fact-checking assistant. Your task is to rank all the given answers from most to least factually correct based on your internal knowledge.\\

Question: \textbf{\{\{question\}\}}

\textbf{\{\{answers\_block\}\}}\\

Please first provide your explanation, and then state your final verdict in the format: `**Final Verdict**: <verdict> AnswerX > AnswerY > AnswerZ </verdict>'. Example: `**Final Verdict**: <verdict> Answer3 > Answer1 > Answer2 </verdict>'.
\end{promptbox}
\end{figure*}

\begin{figure*}[t]
\begin{promptbox}[title={Prompt ~\thetcbcounter: System Prompt for Agentic Generative Verification},
label={prompt:generative_verification}]
Based on the preceding conversation, your task is to rank all the given answers from most to least factually correct.\\

Question: \textbf{\{\{question\}\}}

\textbf{\{\{answers\_block\}\}}\\

Please first provide your explanation, and then state your final verdict in the format: `**Final Verdict**: <verdict> AnswerX > AnswerY > AnswerZ </verdict>'. Example: `**Final Verdict**: <verdict> Answer3 > Answer1 > Answer2 </verdict>'.
\end{promptbox}
\end{figure*}

\begin{figure*}[t]
\begin{promptbox}[title={Prompt ~\thetcbcounter: System Prompt for Agentic Generative Scoring},
label={prompt:generative_scoring}]
Based on the preceding conversation, your task is to score the factuality of the given answer from a scale of 1-10.\\

Question: \textbf{\{\{question\}\}}

Answer: \textbf{\{\{answer\}\}}\\

Please first provide your explanation, and then state your final verdict in the format: `**Final Verdict**: <verdict> a score between 1-10 </verdict>'. Example: `**Final Verdict**: <verdict> 3 </verdict>'.
\end{promptbox}
\end{figure*}









\label{sec:appendix}
\end{document}